\documentclass[lettersize,journal]{IEEEtran}

\usepackage[letterpaper, top=3.2cm, bottom=3.2cm, left=1.92cm, right=1.92cm]{geometry}
\usepackage{multirow}
\usepackage[table,xcdraw]{xcolor}
\usepackage{graphicx}
\usepackage{amsmath}
\usepackage{amsfonts}
\usepackage{algorithm}
\usepackage{amsthm,amssymb}
\usepackage{mathrsfs}
\usepackage{algorithmic}
\usepackage{makecell}
\usepackage{easyReview}
\usepackage{booktabs}
\setcellgapes{3pt}
\usepackage{verbatim}
\setcellgapes{3pt}

\usepackage[justification=centering]{caption}
\usepackage{color}
\usepackage[subfigure]{tocloft}
\usepackage{subfigure}
\usepackage{tabularx}
\usepackage{amsmath, bm}
\usepackage{hyperref}
\hypersetup{hidelinks,
	colorlinks=true,
	allcolors=blue,
	pdfstartview=Fit,
	breaklinks=true}

\usepackage{cite}

\ifCLASSINFOpdf
\else
\fi
\begin{document}
	\title{SIMAC: A Semantic-Driven Integrated Multimodal Sensing And Communication Framework}
	
	\author{Yubo Peng, \textit{Student Member, IEEE}, Luping Xiang, \textit{Senior Member, IEEE}, Kun Yang, \textit{Fellow, IEEE}, Feibo Jiang, \textit{Senior Member, IEEE}, Kezhi Wang, \textit{Senior Member, IEEE}, and Dapeng Oliver Wu, \textit{Fellow, IEEE}
		\thanks{
			{Yubo Peng (ybpeng@smail.nju.edu.cn), Luping Xiang (luping.xiang@nju.edu.cn), and Kun Yang (kunyang@essex.ac.uk) are with the State Key Laboratory of Novel Software Technology, Nanjing University, Nanjing, China, and the School of Intelligent Software and Engineering, Nanjing University (Suzhou Campus), Suzhou, China.
    
            Feibo Jiang (jiangfb@hunnu.edu.cn) is with the School of Information Science and Engineering, Hunan Normal University, Changsha, China. 

            Kezhi Wang (Kezhi.Wang@brunel.ac.uk) is with the Department of Computer Science, Brunel University London, UK.

            Dapeng Oliver Wu (dpwu@ieee.org) is with the Department of Computer Science, City University of Hong Kong, Hong Kong.
			}
		}
	}

\markboth{Submitted for Review}%
{Shell \MakeLowercase{\textit{et al.}}: Bare Demo of IEEEtran.cls for IEEE Journals}

\maketitle 

\begin{abstract}
Traditional single-modality sensing faces limitations in accuracy and capability, and its decoupled implementation with communication systems increases latency in bandwidth-constrained environments. Additionally, single-task-oriented sensing systems fail to address users' diverse demands.
To overcome these challenges, we propose a semantic-driven integrated multimodal sensing and communication (SIMAC) framework. This framework leverages a joint source-channel coding architecture to achieve simultaneous sensing decoding and transmission of sensing results.
Specifically, SIMAC first introduces a multimodal semantic fusion (MSF) network, which employs two extractors to extract semantic information from radar signals and images, respectively. MSF then applies cross-attention mechanisms to fuse these unimodal features and generate multimodal semantic representations.
Secondly, we present a large language model (LLM)-based semantic encoder (LSE), where relevant communication parameters and multimodal semantics are mapped into a unified latent space and input to the LLM, enabling channel-adaptive semantic encoding.
Thirdly, a task-oriented sensing semantic decoder (SSD) is proposed, in which different decoded heads are designed according to the specific needs of tasks. Simultaneously, a multi-task learning strategy is introduced to train the SIMAC framework, achieving diverse sensing services.
Finally, experimental simulations demonstrate that the proposed framework achieves diverse sensing services and higher accuracy.
\end{abstract}


%
\IEEEpeerreviewmaketitle

\section{Introduction}
\subsection{Backgrounds}
Due to distinctive advantages, single-modality sensing technologies, such as radar and visual sensing, have been extensively studied and widely applied across various domains. In autonomous driving, radar sensing facilitates precise object distance, velocity, and relative motion measurements, making it indispensable for advanced vehicular systems \cite{sun2020mimo}. In the military sector, radar sensing is a cornerstone of target reconnaissance, surveillance, missile early warning, and anti-missile defense systems \cite{aravinda2024iot}. 
On the other hand, visual sensing excels in capturing rich image data, enabling detailed recognition and classification of objects in diverse environments. Companies such as Tesla and Waymo have adopted sophisticated visual perception algorithms in their autonomous driving platforms to enhance decision-making and situational awareness \cite{chen2022milestones}. In intelligent surveillance systems, visual perception is employed for automated video analysis, including intruder detection, anomaly identification, and fire monitoring, thereby significantly enhancing security and situational management \cite{nawaratne2019spatiotemporal}.  

Despite their advantages, both modalities have inherent limitations. Radar sensing, while adept at providing physical location and movement data, cannot offer visual details such as color, texture, or shape. Conversely, visual sensing faces challenges in accurately determining spatial positions, particularly in complex or dynamic environments, and is highly sensitive to environmental factors such as lighting and occlusion \cite{wang2024multi}.  
To overcome these limitations, multimodal sensing has emerged as a promising solution, integrating radar and visual sensing to capitalize on the strengths of both modalities \cite{liu2023multimodal}. By combining the spatial information provided by radar with the detailed visual data captured by cameras, multimodal sensing delivers a more comprehensive understanding of the environment. 

However, the traditional decoupled architecture of sensing and communication significantly limits the potential of multimodal sensing. In such systems, the sensing task is completed at the transmitter before transmitting the sensing results to the receiver \cite{10403776}. This separation creates two key issues: first, the sequential process increases service latency, which is particularly problematic for multimodal systems requiring real-time responses; second, the inherently larger data volume generated by multimodal sensing results in substantial communication overhead. These limitations not only reduce the efficiency of multimodal sensing, but also hinder its deployment in scenes with strict latency and bandwidth constraints.

\subsection{Challenges}
Given this background, several challenges associated with traditional sensing technology are summarized as follows:
\begin{enumerate}
    \item \textit{Insufficient Information Sensing:}
    Each modality has inherent limitations and cannot provide richer sensing information. For example, radar sensing, while effective at determining the physical location and movement of targets, cannot provide visual details, such as color, texture, or shape. Conversely, visual sensing struggles to accurately determine the spatial location of objects, particularly in complex or dynamic scenes, and is highly susceptible to environmental conditions like lighting and occlusion.  

    \item \textit{High Communication Overhead:}
    The traditional decoupled architecture of sensing and communication limits the efficiency of multimodal systems. By completing sensing at the transmitter and then transmitting results to the receiver, this approach generates high communication overhead due to the large data volumes inherent in multimodal sensing, particularly in bandwidth-constrained environments. This separation significantly hinders the practical deployment of multimodal systems in real-time applications.

    \item \textit{Limited Sensing Services:}
    On the one hand, relying solely on unimodal sensing significantly limits the ability to provide enriched sensing content. For instance, it is challenging to obtain detailed imagery of a target using only radar-based sensing methods. On the other hand, most sensing algorithms are typically optimized for specific tasks, such as measuring distance, estimating velocity, or capturing imagery. Consequently, it becomes difficult to simultaneously cater to the diverse or personalized sensing needs of users.
\end{enumerate}

\subsection{Contributions}
Semantic communication (SC) is an innovative approach based on deep joint source-channel coding (JSCC), with the potential to transform communication system design and development \cite{10558819,10670195}. Unlike conventional systems, SC focuses on understanding and conveying the message's core meaning or intent, rather than transmitting all bits \cite{10679559}. This paradigm shift allows for reduced redundancy and irrelevant data, improving transmission efficiency.

Integrated sensing and communication (ISAC) unifies sensing and data transmission, overcoming the limitations of traditional decoupled systems \cite{luo2024optimizing}. By enabling simultaneous sensing and communication, ISAC reduces latency and communication overhead, making it particularly effective for multimodal systems. This approach optimizes resource use and enhances efficiency in real-time applications.

Based on SC and ISAC, a semantic-driven integrated multimodal sensing and communication (SIMAC) framework is proposed to address the identified challenges. The main contributions are as follows:
\begin{enumerate}
    \item We introduce a multimodal semantic fusion (MSF) network that employs two extractors to extract unimodal semantics from images and radar signals, respectively. One is based on the vision transformer (ViT) and the other is based on complex convolutional neural networks (CNNs). Then a cross-attention mechanism is used to fuse these unimodal semantics, obtaining a comprehensive multimodal semantic representation. This approach fully leverages both physical position and visual information, addressing the first challenge.

    \item We present an LLM-based semantic encoder (LSE), where a specialized embedding network maps both multimodal semantics and relevant communication parameters into a unified latent space and obtains an embedding. Then, an LLM is applied to perform semantic encoding on the generated embeddings. Due to the inclusion of communication parameters, LSE can flexibly adapt to various communication environments without retraining. Compared to traditional methods, only the semantic encoding needs transmission, reducing communication overheads and addressing the second challenge.

    \item We design a multi-task-oriented sensing semantic decoder (SSD) with distinct decoding heads tailored to specific tasks, such as distance and angle prediction, velocity estimation, and image reconstruction. Additionally, a multi-task learning strategy is implemented to train these heads simultaneously, enhancing training efficiency. This approach enables users to access diverse sensing services, addressing the final challenge.

    \item Based on the VIRAT Video Dataset \cite{oh2011large}, we construct a specific dataset to train and evaluate the proposed framework. The results demonstrate that our framework provides more diverse sensing services and higher accuracy with low communication costs.
\end{enumerate}

\subsection{Organization}
The rest of the paper has the following structure: Section II introduces the related works, and Section III provides a detailed description of the system model. 
Section IV presents the proposed SIMAC framework, including the implementation of the MSF, LSE, and SSD modules. 
Section V employs experimental simulations to evaluate the performance of the proposed methods. 
Lastly, Section VI concludes this paper.

\section{Related Works}
This section reviews the related works about unimodal and multimodal sensing and ISAC. We also summarize the differences between the existing works and ours in Table \ref{tab:compare}.

\renewcommand{\arraystretch}{1.5} 
\begin{table*}[htbp]
	\centering
	\caption{Comparison of our contributions with related literature}
	\label{tab:compare}
	\begin{tabular}{|c|c|c|c|c|c|c|c|c|c|c|}
\hline
Contributions          & Ours                      & \cite{9429942} & \cite{sohail2023radar} & \cite{luo2023computer} & \cite{10064181} & \cite{deliali2023framework} & \cite{kim2023craft} & \cite{10437603} & \cite{10791445} & \cite{10313997} \\ \hline
Unimodal sensing       & \checkmark & \checkmark       & \checkmark               & \checkmark               & \checkmark        & \checkmark                    & \checkmark            & \checkmark                    & \checkmark        & \checkmark        \\ \hline
Multimodal sensing     & \checkmark &                                 &                                         &                                         & \checkmark        & \checkmark                    & \checkmark            &                                              &                                  &                                  \\ \hline
ISAC                   & \checkmark &                                 &                                         &                                         &                                  &                                              &                                      & \checkmark                    & \checkmark        & \checkmark        \\ \hline
Semantic communication & \checkmark &                                 &                                         &                                         &                                  &                                              &                                      &                                              &                                  &                                  \\ \hline
Channel self-adaption  & \checkmark &                                 &                                         &                                         & \checkmark        &                                              &                                      &                                              & \checkmark        &                                  \\ \hline
Multi-task learning    & \checkmark &                                 &                                         &                                         &                                  &                                              &                                      &                                              &                                  &                                  \\ \hline
\end{tabular}
\end{table*}

\subsection{Unimodal Sensing}  
Sensing technologies have found extensive applications across various fields, due to their effectiveness in target detection and tracking, particularly in complex environments. For instance, in autonomous driving, Sun \textit{et al.} \cite{9429942} developed a high-resolution imaging radar system that delivers high-fidelity four-dimensional (4D) sensing through joint sparsity optimization in the frequency spectrum and array configurations. Sohail \textit{et al.} \cite{sohail2023radar} proposed a radar-based method for relative vehicle positioning, utilizing the dynamic range and azimuth of frequency-modulated continuous wave radar to achieve precise vehicle positioning. Additionally, Luo \textit{et al.} \cite{luo2023computer} applied computer vision (CV)-based surface defect detection to monitor the status and integrity of bridge structures, ensuring their safety and reliability.  

While these studies exploit the advantages of sensing technologies, they rely on unimodal data, which inherently limits the diversity and scope of their sensing capabilities. \textit{Therefore, we propose an approach that integrates radar signals with visual data. In this method, the radar signals assist in locating the key target in rough visual information, while the visual modality improves the accuracy of the motion parameters estimation.}

\subsection{Multimodal Sensing}  
The limitations of single-modal sensing, such as challenges in maintaining robustness and accuracy in complex environments, have driven interest in multimodal sensing technologies. Liu and Lin \cite{10064181} introduced a multimodal dynamic hand gesture recognition method using a two-branch deformable network with Gram matching, ensuring reliable recognition and improving generalization across varying field-of-view scenes. Deliali \textit{et al.} \cite{deliali2023framework} developed a framework to classify radar-based trajectories in multimodal traffic environments, enhancing performance under adverse lighting and weather conditions. Kim \textit{et al.} \cite{kim2023craft} proposed an early fusion method that combines spatial and contextual properties from cameras and radar for improved 3D object detection.  

Although these works demonstrate advancements in multimodal fusion and sensing accuracy, they largely overlook the communication cost of transmitting sensing results. Furthermore, their designs are often tailored to specific tasks, such as detection or imaging. \textit{In this paper, we will address these gaps by introducing LSE to mitigate dynamic communication environment challenges and employing SSD to efficiently deliver diverse sensing services to users.}  

\subsection{Integrated Sensing and Communication}  
ISAC offers a paradigm shift by unifying sensing and communication into a single framework, significantly improving system efficiency. Zhang \textit{et al.} \cite{10437603} investigated an IRS-assisted, WPT-enabled ISAC system, where a base station (BS) performs both radar sensing and data reception from IoT devices. To extend ISAC's potential in cellular-connected UAV systems, Wang \textit{et al.} \cite{10791445} proposed an extended Kalman filtering-based data fusion algorithm, providing precise environmental information and enabling beyond-line-of-sight (LoS) sensing. Xiang \textit{et al.} \cite{10313997} developed a green beamforming design for ISAC, employing beam-matching error to evaluate radar performance.  

While these studies achieve parallel data transmission and sensing, they fall short in two critical areas. First, they lack multimodal perception capabilities. Second, they transmit raw data, resulting in substantial communication overhead. \textit{In contrast, our approach combines MSF for effective multimodal fusion with SC to minimize communication costs, significantly enhancing the system's overall efficiency and scalability.}

More importantly, traditional ISAC systems typically achieve synesthetic parallelism by transmitting sensing signals alongside communication data. However, this approach introduces significant communication overhead and necessitates a trade-off between sensing and communication performance. Moreover, it is highly susceptible to interference from channel noise.  
\textit{Our proposed SIMAC framework leverages a semantic-driven approach to synesthetic integration, deeply embedding sensing information into communication semantics and transmitting only the essential semantic content. This innovative design effectively addresses the limitations of traditional methods, reducing communication overhead and enhancing robustness against channel noise.} 

\begin{figure*}[htbp]
	\centering
	\includegraphics[width=16cm]{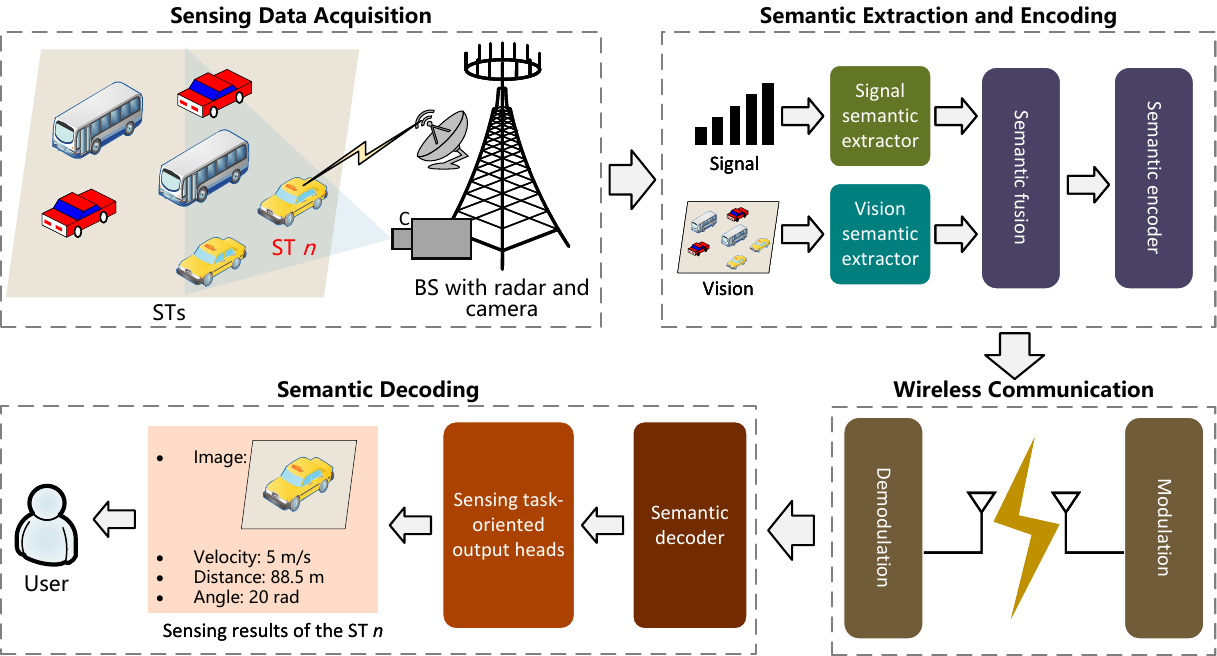}
	\caption{The illustration of the integrated SC and multimodal sensing system model.}
	\label{fig:sys}
\end{figure*}
\section{Wireless Sensing and Communication System Model}
As illustrated in Fig. \ref{fig:sys}, we consider a system consisting of $N$ sensing targets (STs), a BS as the transmitter, and a user as the receiver. The primary objective of the BS is to transmit the image of the $n$th ST (i.e., communicated data) and its motion parameters (i.e., sensing results) to the user. Specifically, the BS utilizes its radar and camera to sense the $n$th ST from both digital signal and visual perspectives. The BS then uses an integrated SC and multimodal sensing system to communicate with the user, where the sensing decoding and data transmission are processed parallelly. The detailed process is outlined as follows:
\subsubsection{Sensing Data Acquisition}
\begin{figure}[htbp]
	\centering
	\includegraphics[width=8cm]{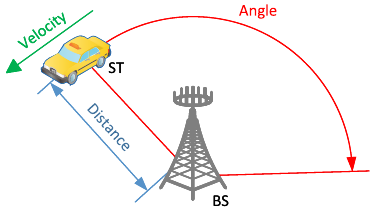}
	\caption{The illustration of the motion parameters of the ST.}
	\label{fig:motion}
\end{figure}
Due to its favorable properties, the linear frequency modulation (LFM) waveform is extensively employed in radar sensing. The frequency of this waveform varies linearly over time, making it inherently robust against doppler frequency shifts, which enhances signal processing gain. Assuming that the motion parameters of the ST \( n \) include the angle \( \theta_{n} \), distance \( d_{n} \), and radial velocity \( v_{n} \), as shown in Fig. \ref{fig:motion}, we adopt a single-input-multiple-output (SIMO) radar model to transmit the LFM waveform and capture the corresponding echo signal. The echo signal can be expressed as: 
\begin{equation}\label{eq:sig}
    \mathbf{A}_{n} = \lambda \mathbf{a}(\theta_{n})\text{e}^{j2\pi\mu_{n}\mathbf{T}}s(\mathbf{T}-\tau_{n}),
\end{equation}
where,
\begin{equation}
   \lambda = \frac{\xi \cdot \rho_n}{(4\pi)^{3/2} \cdot d_n^2},
\end{equation}
and
\begin{equation}
    \xi = \frac{c}{f_c + K_t},
\end{equation}
while $\rho_n$ is the radar cross-section (RCS) of the $n$th ST, which represents the ST's ability to reflect radar signals. \( \mathbf{a}(\theta_{n}) = [1, \text{e}^{-j\pi\cos\theta_{n}}, \ldots, \text{e}^{-j(K-1)\pi\cos\theta_{n}}] \) is the steering vector, and \( K \) represents the number of receive antennas. \( \mathbf{T} = [k \cdot \Delta t \mid k \in \mathbb{Z}, 0 \leq k \Delta t \leq T_r ] \) denotes the time sequence of the sampling process, where \( T_r \) is the pulse repetition interval (PRI), \( \Delta t = \frac{1}{F_s} \) is the sampling interval, and \( F_s \) is the sampling frequency.  
\( \tau_{n} = \frac{2d_{n}}{c} \) and \( \mu_{n} = \frac{2(f_c+K_t/2) v_{n}}{c} \) represent the time delay and doppler frequency shift, respectively, where \( f_c \) is the radar's central frequency, $K_t$ is the frequency modulation slope, and \( c = 3 \times 10^8 \, \text{m/s} \) is the velocity of light.

BGR mode represents image pixels using blue, green, and red values, allowing direct display of sensor data without interpolation for optimal quality. Thus, it is widely used in image sensors \cite{tan2024integrating}.
We assume that the BS is equipped with a camera that utilizes the BGR mode to acquire high-quality images $\mathbf{m} \in \mathbb{R}^{W \times H \times 3}$, where $W$ and $H$ denote the width and height of the image in terms of the number of pixels, respectively. Note the captured image $\mathbf{m}$ may contain multiple STs, hence we aim to isolate the portion of the image corresponding to the $n$th ST, denoted as $\mathbf{m}_{n}$, using the latent information of the echo signal $\mathbf{A}_{n}$.

\subsubsection{Semantic Extraction}
Given that the echo signal $\mathbf{A}_{n}$ and the captured image $\mathbf{m}$ have distinct data dimensions and characteristics, we adopt two separate semantic extractors for each modality. The process of semantic extraction is described as follows:
\begin{equation}
    \mathbf{s}_{n}^\text{sig}=S_\text{sig}(\mathbf{A}_{n}, \bm{\alpha}),
\end{equation}
\begin{equation}
    \mathbf{s}_{n}^\text{vis}=S_\text{vis}(\mathbf{m}, \bm{\beta}),
\end{equation}
where $\mathbf{s}_{n}^\text{sig} $ and $\mathbf{s}_{n}^\text{vis}$ represent the semantic features of length $L_s$ extracted from $\mathbf{A}_{n}$ and $\mathbf{m}$, respectively. $S_\text{sig}(\cdot)$ denotes the signal semantic extractor with parameters $\bm{\alpha}$, and $S_\text{vis}(\cdot)$ is the image semantic extractor with parameters $\bm{\beta}$. 

To efficiently capture key information and explore the latent relationships between the two semantic features $\mathbf{s}_{n}^\text{sig}$ and $\mathbf{s}_{n}^\text{vis}$, a semantic fusion module is employed to combine these features and generate a comprehensive multimodal semantic representation $\mathbf{s}_{n}^\text{mul}$. This process can be expressed as:
\begin{equation}
    \mathbf{s}_{n}^\text{mul}=S_\text{mul}(\mathbf{s}_{n}^\text{sig}, \mathbf{s}_{n}^\text{vis}, \bm{\gamma}),
\end{equation}
where $S_\text{mul}(\cdot)$ is the semantic fusion module with parameters $\bm{\gamma}$.

\subsubsection{Semantic Encoding}
To minimize semantic distortion during wireless transmission, a JSCC encoder is employed to perform semantic encoding, taking $\mathbf{s}_{n}^\text{mul}$ as input. The encoding process is given by:
\begin{equation}\label{eq:se}
    \mathbf{e}_{n} = F_\text{se}(\mathbf{s}_{n}^\text{mul},\bm{\delta}),
\end{equation}
where $\mathbf{e}_{n}$ represents the semantic encoding and $F_\text{se}(\cdot)$ is the JSCC encoder with parameters $\bm{\delta}$. 

\subsubsection{Wireless Communication}
To ensure that the semantic encoding can be transmitted over the wireless channel, signal modulation techniques, such as QPSK and 16QAM, are employed to convert $\mathbf{e}_{n}$ into complex-valued symbols $\mathbf{c}_{n}$. 

The complex-valued symbols $\mathbf{c}_{n}$ are transmitted over the channel, which is modeled as:
\begin{equation}\label{eq:trans}
	\mathbf{y}_{n} = \mathbf{H} \cdot \mathbf{c}_{n} + \mathbf{N},
\end{equation}
where $\mathbf{y}_{n}$ represents the received complex-valued symbols, $\mathbf{H}$ denotes the channel gain between the user and the BS, and $\mathbf{N}$ represents Additive White Gaussian Noise (AWGN). Given that we consider a deep JSCC architecture, the channel model must be compatible with backpropagation to facilitate end-to-end training of both the encoder and decoder. Consequently, the wireless channel is simulated using neural network-based approaches \cite{jiang2024large}.

During wireless communication between the BS and the user, the transmission rate can be expressed as:
\begin{equation}
	v = B \log _{2} \left( 1 + \frac{P \mathbf{H}}{\mathbf{N}} \right),
\end{equation}
where $B$ and $P$ represent the bandwidth and transmission power, respectively, when the BS communicates with the user. Thus, the transmission delay is given by:
\begin{equation}
	t^\text{com} = \frac{Z(\mathbf{c}_{n})}{v},
\end{equation}
where $Z(\mathbf{c}_{n})$ denotes the number of bits required to transmit the complex-valued symbols $\mathbf{c}_{n}$ to the user.

\subsubsection{Semantic Decoding}
Upon receiving the symbols $\mathbf{y}_{n}$, signal demodulation is applied to convert them back into the received semantic encoding $\hat{\mathbf{e}}_{n}$. To support diverse sensing services for the user, a JSCC decoder is employed to obtain sensing results tailored for multiple tasks. Specifically, we consider a variety of sensing tasks, including distance, angle, velocity prediction, and ST reconstruction. The decoding process is therefore formulated as:
\begin{equation}
    \textbf{o}_{n} = F_\text{sd}(\hat{\mathbf{e}}_{n}, \bm{\epsilon}),
	\quad \textbf{o}_{n} \in \{\hat{\theta}_{n}, \hat{d}_{n}, \hat{v}_{n}, \hat{\mathbf{m}}_{n}\},
\end{equation}
where $F_\text{sd}(\cdot)$ represents the semantic decoder parameterized by $\bm{\epsilon}$. The decoded results, $\textbf{o}_{n}$, may include a selection of the reconstructed image $\hat{\mathbf{m}}_{n}$ of the $n$th ST, predicted distance $\hat{d}_{n}$, predicted velocity $\hat{v}_n$, and estimated angle $\hat{\theta}_{n}$.

\subsection{Problem Formulation}
To finish the data transmission and sensing decoding from the BS to the user via a wireless channel, the total execution time $T^{\mathrm{exe}}$ comprises the computation time  for semantic extraction $t^{\mathrm{st}}$ and encoding $t^{\mathrm{se}}$ at the transmitter, the communication time for transmission $t^{\mathrm{com}}$, and the computation time for semantic decoding at the receiver $t^{\mathrm{sd}}$. Thus, the total execution time can be expressed as:
\begin{equation}
T^{\mathrm{exe}} = t^{\mathrm{st}} + t^{\mathrm{se}} + t^{\mathrm{com}} + t^{\mathrm{sd}}.
\end{equation}

To provide diversified services for the user, we consider multiple sensing tasks, including distance, velocity, angle prediction, and image reconstruction of the $n$th ST. The corresponding task losses are defined as follows:
\begin{equation}\label{eq:loss1}
    \mathcal{L}_\text{dp} = ||d_{n} - \hat{d}_{n}||^2,
\end{equation}
\begin{equation}\label{eq:loss2}
    \mathcal{L}_\text{ap} = ||\theta_{n} - \hat{\theta}_{n}||^2,
\end{equation}
\begin{equation}\label{eq:loss3}
    \mathcal{L}_\text{vp} = ||v_{n} - \hat{v}_{n}||^2,
\end{equation}
\begin{equation}\label{eq:loss4}
    \mathcal{L}_\text{sr} = ||\mathbf{m}_{n} - \hat{\mathbf{m}}_{n}||.
\end{equation}

The primary goal of the SIMAC framework is to minimize semantic distortion during wireless transmission while maximizing the accuracy of the decoded sensing results. Additionally, transmission delays must be accounted for to ensure the quality of service. Accordingly, the objective function of the proposed SIMAC framework can be expressed as:  
\begin{subequations}\label{eq:problem}  
\begin{align}  
\min_{\bm{\alpha}, \bm{\beta}, \bm{\gamma}, \bm{\delta}, \bm{\epsilon}} l_1\mathcal{L}_\text{dp} + l_2\mathcal{L}_\text{ap} + l_3\mathcal{L}_\text{vp} + l_4\mathcal{L}_\text{sr},  
\end{align}  
\begin{alignat}{1}  
\text{s.t.~}  
& T^\text{exe} \leq T_\text{max},  
\end{alignat}  
\end{subequations}  
where $T_\text{max}$ denotes the latency requirement for completing the sensing task. $l1$, $l2$, $l3$, and $l4$ are adjustment factors.

To address the optimization problem described in Eq. (\ref{eq:problem}a), there are three key issues. First, images and radar signals have different modalities and are difficult to process using a single neural network. Second, fixed-strategy semantic encoding is difficult to adapt to dynamic changes in communication parameters, resulting in low semantic fidelity. Finally, traditional single-task learning has difficulty optimizing multiple objectives simultaneously.
Therefore, we have meticulously designed specialized neural networks for the various modules within the SIMAC framework, which will be described in the next section.

\section{Semantic-Driven Integrated Multimodal Sensing and Communication}
\begin{figure*}[htbp]
	\centering
	\includegraphics[width=16cm]{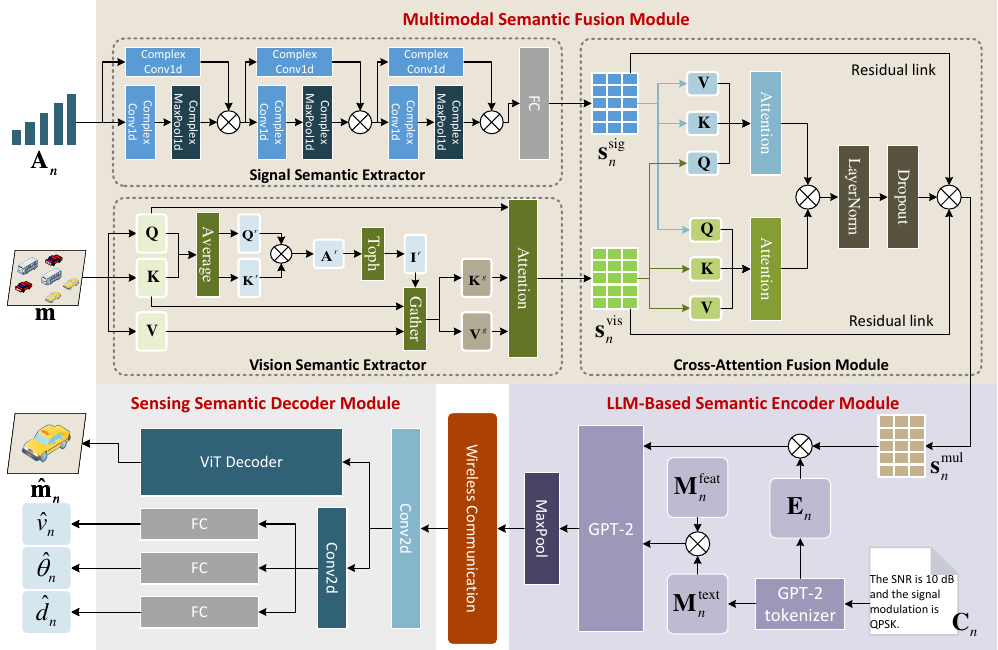}
	\caption{The network design of the proposed SIMAC framework.}
	\label{fig:SIMAC}
\end{figure*}
\subsection{Overview}
Single-modal sensing is often insufficient to meet the demands of diverse and highly accurate sensing services. Furthermore, traditional sensing and communication systems are typically designed in a decoupled manner, which exacerbates service latency, particularly in real-time applications \cite{10077112}. 
Therefore, we propose the SIMAC framework, which integrates multimodal sensing with SC to overcome the limitations of conventional sensing and communication systems. As depicted in Fig. \ref{fig:SIMAC}, the SIMAC framework consists of three primary modules:   

\subsubsection{MSF for Multimodal Semantic Representation}
Given the heterogeneity of multimodal data, MSF leverages complex CNNs and ViT as semantic extractors, \( S_\text{sig}(\cdot) \) and \( S_\text{vis}(\cdot) \), for processing signal and visual modalities, respectively. A fusion network \( S_\text{mul}(\cdot) \), based on a cross-attention mechanism, is then employed to perform deep semantic fusion, resulting in a multimodal semantic representation, \( \mathbf{s}_{n}^\text{mul} \). Details on the design of the MSF module are provided in Section IV-B.  
Theoretically, the semantic representation encapsulates the critical information inherent in the original multimodal data, making it suitable as the source input for semantic communication.  

\subsubsection{LSE for Channel-Adaptive Semantic Encoding}
In the traditional communication system, some communication parameters, such as the signal modulation scheme and the signal-to-noise ratio (SNR), are dynamic. Therefore, LSE is employed as the JSCC encoder, benefiting from its ability to consider these communication parameters while performing semantic encoding. Hence, Eq. (\ref{eq:se}) can be rewritten as:
\begin{equation}
    \mathbf{e}_{n} = F_\text{se}(\mathbf{s}_{n}^\text{mul}, C_n,\bm{\delta}),
\end{equation}
where $C_n$ represents the communication parameters that are in the format of natural languages, for example, ``the SNR is 5 dB and the signal modulation is QPSK."
We describe the details of LSE in Section IV-C.
As a result, the semantic encoding $\mathbf{e}_{n}$ can flexibly adapt to various communication environments by adjusting $C_n$ accordingly. Thereafter, the semantic encoding is modulated as the complex-valued symbols $\mathbf{c}_{n}$ and transmitted on the physical channel.

\subsubsection{SSD for Task-Oriented Semantic Deconding}
Upon receiving noisy symbols \( \mathbf{y}_{n} \) at the receiver, signal demodulation is performed to recover the received semantic encoding \( \hat{\mathbf{e}}_{n} \). To support diversified sensing services for the user, the task-oriented SSD is employed as the JSCC decoder to generate task-specific sensing results. The SSD is designed with multiple output heads to handle a variety of tasks, including distance estimation, velocity estimation, angle prediction, and image reconstruction of the $n$th ST. Further details on the SSD module are provided in Section IV-D.  

Multi-task learning is adopted during training to ensure the accurate execution of different sensing tasks, as will be described in Section IV-E.  

Assuming that the training dataset is $\mathcal{D}$, the workflow of the proposed SIMAC framework is outlined in \textbf{Algorithm \ref{alg:SIMAC}}. Unlike traditional ISAC systems, the SIMAC framework adopts a semantic-driven approach to synesthetic integration by deeply embedding sensing information within communication semantics and transmitting only the essential semantic content. This innovative design effectively overcomes the limitations of conventional methods, where the communication overhead is reduced significantly and the robustness against noise is drastically improved. 

\begin{algorithm}
    \caption{SIMAC Framework Workflow}
    \label{alg:SIMAC}
    \begin{algorithmic}[1]
        \REQUIRE \(\mathbf{m}, \mathbf{A}_{n}, \mathcal{D}\).
        \ENSURE \(\textbf{o}_{n}, \bm{\alpha}, \bm{\beta}, \bm{\gamma}, \bm{\delta}, \bm{\epsilon}\)  

        \vspace{0.5em}
        \noindent\textbf{Inference Phase:}
        \STATE Obtain the semantic representation \(\mathbf{s}_{n}^\text{mul}\) from multimodal data \(\mathbf{m}\) and \(\mathbf{A}_{n}\) using \textbf{Algorithm \ref{alg:MSF}}.
        \STATE Obtain the channel-adaptive semantic encoding \(\mathbf{e}_n\) using \textbf{Algorithm \ref{alg:LSE}}.
        \STATE Modulate \(\mathbf{e}_{n}\) into \(\mathbf{c}_{n}\) and perform wireless transmission according to Eq. (\ref{eq:trans}).
        \STATE Perform task-oriented semantic decoding using \textbf{Algorithm \ref{alg:SSD}} and obtain the sensing result \(\textbf{o}_{n}\).

        \vspace{0.5em}
        \noindent\textbf{Training Phase:}
        \STATE Obtain the trained parameters \(\bm{\alpha}, \bm{\beta}, \bm{\gamma}, \bm{\delta}, \bm{\epsilon}\) by jointly training all the modules according to \textbf{Algorithm \ref{alg:MTL}}, using \(\mathcal{D}\).
    \end{algorithmic}
\end{algorithm}

\subsection{Multimodal Semantic Fusion Module}
MSF integrates radar signal and visual image modalities through a carefully designed framework that combines signal processing, transformer-based image feature extraction, and cross-attention fusion, as shown in Fig. \ref{fig:SIMAC}. The detailed description of key modules is as follows:

\subsubsection{Signal Semantic Extractor}
We first process radar signal $\mathbf{A}_{n}$, which is represented as complex-valued tensors of shape \((B, K, L_\text{sig})\), where \(B\) is the batch size, \(K\) is the number of receiving antennas used by the BS for sensing, and \(L_\text{sig}\) is the input length. Then, each complex-valued convolutional layer performs operations on both the real and imaginary components of the signal $\mathbf{A}_{n}$, denoted by $\mathbf{x}_{n}^\text{real}$ and $\mathbf{x}_{n}^\text{imag}$, respectively. This process can be expressed as:  

\begin{equation}\label{eq:sse2}
    \mathbf{z}_{n}^\text{real} = \mathbf{W}_\text{real} \ast \mathbf{x}_{n}^\text{real} - \mathbf{W}_\text{imag} \ast \mathbf{x}_{n}^\text{imag},
\end{equation}
\begin{equation}\label{eq:sse3}
    \mathbf{z}_{n}^\text{imag} = \mathbf{W}_\text{real} \ast \mathbf{x}_{n}^\text{imag} + \mathbf{W}_\text{imag} \ast \mathbf{x}_{n}^\text{real},
\end{equation}
\begin{equation}\label{eq:sse1}
    \mathbf{z}_{n}^\text{out} = \text{ReLU}(\mathbf{z}_{n}^\text{real} + j \cdot \mathbf{z}_{n}^\text{imag}),
\end{equation}
where \(\ast\) denotes convolution, \(\mathbf{W}_\text{real}\) and \(\mathbf{W}_\text{imag}\) are the real and imaginary parts of the convolutional kernel, and \(j\) represents the imaginary unit. Next, complex max-pooling is applied, reducing the sequence length while retaining the semantic features. After three convolutional layers, each followed by a pooling operation, the real and imaginary parts are concatenated along the channel dimension:
\begin{equation}\label{eq:sse4}
\mathbf{z}_{n,3} = \text{Concat}(\mathbf{z}_{n,3}^\text{real}, \mathbf{z}_{n,3}^\text{imag}).
\end{equation}

Finally, the signal is sent through by a fully connected layer that maps the concatenated features to a reduced dimensionality:
\begin{equation}\label{eq:sse5}
\mathbf{s}_{n}^\text{sig} = \operatorname{Linear}(\mathbf{z}_{n,3}), \mathbf{s}_{n}^\text{sig} \in  \mathbb{R}^{B \times L_\text{s} \times d},
\end{equation}
where $\operatorname{Linear}(\cdot)$ represents a fully connected layer, $L_\text{s}$ is the sequence length of the signal semantic and $d$ is the feature dimensionality.

\subsubsection{Vision Semantic Extractor}
To ensure high extraction accuracy and inference velocity, we utilize a lightweight ViT network——bilateral transformer (BiFormer) \cite{zhu2023biformer}, as the backbone to extract visual features from input images $\mathbf{m}$. 
The key advantage of BiFormer over traditional ViT lies in its bilevel routing attention (BRA) mechanism.

First, $\mathbf{m}$ is divided into $S \times S$ non-overlapping regions using a patch embedding layer, with each region containing $HW/S^2$ feature vectors. This process reshapes $\mathbf{m}$ into $\mathbf{m}^r \in \mathbb{R}^{S^2 \times HW/S^2 \times C}$. Linear projections are then applied to obtain query, key, and value tensors, denoted as $\mathbf{Q}$, $\mathbf{K}$, and $\mathbf{V} \in \mathbb{R}^{S^2 \times HW/S^2 \times C}$, respectively: \begin{equation}\label{eq:Bi1}
    \mathbf{Q}=\mathbf{m}^r \mathbf{W}^\text{q}, \quad \mathbf{K}=\mathbf{m}^r \mathbf{W}^\text{k}, \quad \mathbf{V}=\mathbf{m}^r \mathbf{W}^\text{v},
\end{equation}
where $\mathbf{W}^\text{q}$, $\mathbf{W}^\text{k}$, and $\mathbf{W}^\text{v}$ are the projection weights for the query, key, and value, respectively.

Second, region-level queries and keys, $\mathbf{Q}^\text{r}$ and $\mathbf{K}^\text{r} \in \mathbb{R}^{S^2 \times C}$, are computed by averaging the query and key tensors over each region. Using these, we calculate the adjacency matrix $\mathbf{A}^\text{r} \in \mathbb{R}^{S^2 \times S^2}$ to quantify semantic relationships between regions: 
\begin{equation}\label{eq:Bi2}
\mathbf{A}^\text{r}=\mathbf{Q}^\text{r}\left(\mathbf{K}^\text{r}\right)^\top,
\end{equation}
where $\top$ represents the transpose operation.

The adjacency matrix is then pruned by retaining the top-$h$ semantic connections for each region, yielding the routing index matrix $\mathbf{I}^\text{r} \in \mathbb{N}^{S^2 \times h}$: 
\begin{equation}\label{eq:Bi3}
    \mathbf{I}^\text{r} = \operatorname{toph}(\mathbf{A}^\text{r}),
\end{equation}
where $\operatorname{toph}(\cdot)$ is the row-wise top-$h$ selection operator. Hence, the $i$-th row of $\mathbf{I}^\text{r}$ contains the indices of the $h$ most semantically relevant regions for the $i$-th region.

Next, with the region-to-region routing index matrix $\mathbf{I}^\text{r}$, fine-grained token-to-token attention is performed by gathering the corresponding key and value tensors:
\begin{equation}\label{eq:Bi4}
    \mathbf{K}^\text{g}=\operatorname{gather}\left(\mathbf{K}, \mathbf{I}^\text{r}\right), \quad \mathbf{V}^\text{g}=\operatorname{gather}\left(\mathbf{V}, \mathbf{I}^\text{r}\right),
\end{equation}
where $\mathbf{K}^\text{g}$ and $\mathbf{V}^\text{g}$ are the gathered key and value tensors. Attention is then applied to the gathered key-value pairs, producing the output tensor:
\begin{equation}\label{eq:Bi5}
    \mathbf{O}=\operatorname{Attention}\left(\mathbf{Q}, \mathbf{K}^\text{g}, \mathbf{V}^\text{g}\right)+\operatorname{LCE}(\mathbf{V}),
\end{equation}
\begin{equation}\label{eq:Bi6}
    \operatorname{Attention}\left(\mathbf{Q}, \mathbf{K}^\text{g}, \mathbf{V}^\text{g}\right)=\operatorname{softmax}\left(\frac{\mathbf{Q}{\mathbf{K}^\text{g}}^\top}{\sqrt{C}}\right) \mathbf{V}^\text{g},
\end{equation}
where $\sqrt{C}$ is the scaling factor, and $\operatorname{LCE}(\mathbf{V})$ refers to a local context enhancement term \cite{ren2022shunted}.

Finally, a linear projection layer $F_\text{Proj}$ is applied to the output tensor to obtain the vision semantic while ensuring it has the same shape with the signal semantic $\mathbf{s}_{n}^\text{sig}$. This can be expressed as:
\begin{equation}\label{eq:Bi7}
    \mathbf{s}_{n}^\text{vis}=F_\text{Proj}(\mathbf{O}), \mathbf{s}_{n}^\text{vis} \in \mathbb{R}^{B \times L_\text{s} \times d}.
\end{equation}

\subsubsection{Cross-Attention Fusion Module}
To achieve the deep multimodal fusion, we integrate the signal semantic $\mathbf{s}_{n}^\text{sig}$ and vision semantic $\mathbf{s}_{n}^\text{vis}$ through a bidirectional attention mechanism \cite{liu2019bidirectional}. Specifically, $\mathbf{s}_{n}^\text{sig}$ acts as the query while $\mathbf{s}_{n}^\text{vis}$ serves as the key and value, and vice versa. The attention outputs are computed as:
\begin{equation}\label{eq:crt1}
    \mathbf{z}_{n}^\text{vis} = \text{Attention}(\mathbf{W}_q^1 \mathbf{s}_{n}^\text{sig}, \mathbf{W}_k^2 \mathbf{s}_{n}^\text{vis}, \mathbf{W}_v^2 \mathbf{s}_{n}^\text{vis}),
\end{equation}
\begin{equation}\label{eq:crt2}
    \mathbf{z}_{n}^\text{sig} = \text{Attention}(\mathbf{W}_q^2 \mathbf{s}_{n}^\text{vis}, \mathbf{W}_k^1 \mathbf{s}_{n}^\text{sig}, \mathbf{W}_v^1 \mathbf{s}_{n}^\text{sig}),
\end{equation}
where \(\mathbf{W}_q^1, \mathbf{W}_k^1, \mathbf{W}_v^1\) are the query, key, and value weights for radar signals, and \(\mathbf{W}_q^2, \mathbf{W}_k^2, \mathbf{W}_v^2\) are the corresponding weights for image features. The fused output is the sum of the two attention outputs:
\begin{equation}\label{eq:crt3}
\mathbf{z}_{n}^\text{fusion} = \mathbf{z}_{n}^\text{vis} + \mathbf{z}_{n}^\text{sig}.
\end{equation}

To avoid the issue of vanishing gradient, normalization and residual connections are employed to refine the fusion:
\begin{equation}\label{eq:crt4}
\mathbf{s}_{n}^\text{mul} = \text{LayerNorm}(\mathbf{z}_{n}^\text{fusion}) + \mathbf{s}_{n}^\text{sig} + \mathbf{s}_{n}^\text{vis}, \mathbf{s}_{n}^\text{mul} \in \mathbb{R}^{B \times L_\text{s} \times d}.
\end{equation}

The inference process of MSF is summarized in \textbf{Algorithm \ref{alg:MSF}}. Overall, the advantage of MSF is the combination of complex operations, ViT-based feature extraction, and attention-driven fusion, effectively fusing the latent information between radar signal processing and image. 

\begin{algorithm}
\caption{Inference of MSF}
\label{alg:MSF}
\begin{algorithmic}[1]
	\REQUIRE $\mathbf{m}$, $\mathbf{A}_{n}$.
	\ENSURE $\mathbf{s}_{n}^\text{mul}$.
	\STATE{Extract signal semantic $\mathbf{s}_{n}^\text{sig}$ using Eqs. (\ref{eq:sse1})-(\ref{eq:sse5}).}
	\STATE{Extract vision semantic $\mathbf{s}_{n}^\text{vis}$ using Eqs. (\ref{eq:Bi1})-(\ref{eq:Bi7}).}
	\STATE{Obtain the fused multimodal semantic $\mathbf{s}_{n}^\text{mul}$ via cross-attention using Eqs. (\ref{eq:crt1})-(\ref{eq:crt4}).}	
\end{algorithmic}
\end{algorithm}

\subsection{LLM-Based Semantic Encoder Module}
LSE is designed to encode multimodal semantic information by leveraging the representational capabilities of LLMs \cite{10638533}. Specifically, the encoder integrates textual descriptions of communication parameters with multimodal features to generate channel-adaptive semantic encoding $\mathbf{e}_{n}$. 

Firstly, the input to the LSE consists of two components: the multimodal semantic representation $\mathbf{s}_{n}^\text{mul}$ derived from the MSF network and textual communication parameters $C_n$. The textual input $C_n$ undergoes tokenization using the GPT-2 tokenizer \cite{veyseh2021unleash}, producing input token IDs \(\mathbf{C}_{n}^\text{text}\) and the corresponding attention mask \(\mathbf{M}_{n}^\text{text}\). The end-of-sequence, ``\texttt{eos\_token}" symbol, replaces padding tokens to ensure compatibility with the pre-trained GPT-2 model \cite{veyseh2021unleash}. The tokenized input is then mapped to an embedding space using the GPT-2 word embedding layer $\text{Embed}(\cdot)$, which can be expressed as:  
\begin{equation}\label{eq:LSE1}
    \mathbf{E}_n = \text{Embed}(\mathbf{C}_{n}^\text{text}), \mathbf{E}_n \in \mathbb{R}^{B \times L_\text{t} \times d},
\end{equation}
where $L_\text{t}$ is the length of the text embedding.

Secondly, the multimodal semantic representation $\mathbf{s}_{n}^\text{mul}$ and textual embeddings \(\mathbf{E}_n\) are concatenated along the sequence dimension to form a fused input:  
\begin{equation}\label{eq:LSE2}
    \mathbf{F}^\text{input}_{n} = \text{Concat}(\mathbf{s}_{n}^\text{mul}, \mathbf{E}_n), \mathbf{F}^\text{input}_{n} \in \mathbb{R}^{B \times L_\text{fusion} \times d},
\end{equation} 
where $L_\text{fusion} = L_\text{s}+L_\text{t}$. Simultaneously, a fused attention mask is constructed:  
\begin{equation}\label{eq:LSE3}
\mathbf{F}^\text{mask}_{n} = \text{Concat}(\mathbf{M}_{n}^\text{feat}, \mathbf{M}_{n}^\text{text}), \mathbf{F}^\text{mask}_{n} \in \mathbb{R}^{B \times L_\text{fusion}},
\end{equation}
where \(\mathbf{M}_{n}^\text{feat} \in \mathbb{R}^{B \times L_\text{s}}\) is initialized as a matrix of ones.

Thirdly, the fused input \(\mathbf{F}^\text{input}_{n}\) and the fused attention mask \(\mathbf{F}^\text{mask}_{n}\) are passed into the pre-trained GPT-2 model, which generates contextually enriched representations from its final hidden state:
\begin{equation}\label{eq:LSE4}
    \mathbf{S}^\text{fusion}_{n} = \text{GPT2}(\mathbf{F}^\text{input}_{n}, \mathbf{F}^\text{mask}_{n}), \mathbf{S}^\text{fusion}_{n} \in \mathbb{R}^{B \times L_\text{fusion} \times d}.
\end{equation}

To reduce the computational burden and extract the most salient features, a max-pooling operation is applied along the sequence dimension, reducing the dimension by half:
\begin{equation}\label{eq:LSE5}
    \mathbf{S}^\text{pool}_{n} = \text{MaxPool}(\mathbf{S}^\text{fusion}_{n}), \mathbf{S}^\text{pool}_{n} \in \mathbb{R}^{B \times L_\text{fusion} \times d/2}.
\end{equation}

Finally, the pooled output is subsequently activated using a hyperbolic tangent function:
\begin{equation}\label{eq:LSE6}
\mathbf{e}_{n} = \tanh(\mathbf{S}^\text{pool}_{n}), \mathbf{e}_{n} \in \mathbb{R}^{B \times L_\text{fusion} \times d/2}.
\end{equation}

We summarize the inference process of LSE in \textbf{Algorithm \ref{alg:LSE}}. Overall, this enriched semantic encoding $\mathbf{e}_{n}$ captures multimodal contextual dependencies and textual semantics, making it suitable for downstream tasks such as semantic reconstruction and communication. By integrating GPT-2 into the pipeline, the LSE effectively bridges the gap between textual and non-textual modalities, achieving a comprehensive representation of multimodal inputs.
\begin{algorithm}
\caption{Inference of LSE}
\label{alg:LSE}
\begin{algorithmic}[1]
    \REQUIRE $\mathbf{s}_{n}^\text{mul}$, $C_{n}$.
    \ENSURE $\mathbf{e}_{n}$.
    \STATE{Tokenize $C_{n}$ to obtain $\mathbf{C}_n^\text{text}$ and $\mathbf{M}_n^\text{text}$.}
    \STATE{Map $\mathbf{C}_n^\text{text}$ to embedding space $\mathbf{E}_n$ using Eq. (\ref{eq:LSE1}).}
    \STATE{Concatenate $\mathbf{s}_{n}^\text{mul}$ and $\mathbf{E}_n$ to obtain $\mathbf{F}_n^\text{input}$ using Eq. (\ref{eq:LSE2}).}
    \STATE{Initiate $\mathbf{M}_n^\text{feat}$.}
    \STATE{Construct fused attention mask $\mathbf{F}_{n}^\text{mask}$ according to Eq. (\ref{eq:LSE3}).}
    \STATE{Generate fused semantic representation $\mathbf{S}_{n}^\text{fusion}$ using GPT-2 according to Eq. (\ref{eq:LSE4}).}
    \STATE{Apply max-pooling to reduce the dimensions using Eq. (\ref{eq:LSE5}).}
    \STATE{Activate pooled representation to obtain $\mathbf{e}_{n}$ by Eq. (\ref{eq:LSE6}).}
\end{algorithmic}
\end{algorithm}

\subsection{Sensing Semantic Decoder Module}
SSD, used as the JSCC decoder, integrates ViT-based and CNN-based modules to extract and decode semantic information and it is capable of reconstructing target images and estimating angles, velocities, and distances. The specific design of SSD is described as follows:

Firstly, SSD reshapes the received semantic encoding $\mathbf{\hat{e}}_n$ into a spatial representation compatible with convolutional operations. To achieve this, the $\operatorname{Conv2d}(\cdot)$ operation rearranges $\mathbf{\hat{e}}_n$ into a size of $(B \times W_d \times H_d \times d/2)$, where \(H_\text{d} = W_\text{d} = \sqrt{L_\text{fusion}}\), assuming \(L_\text{fusion}\) is a perfect square. This process can be expressed as:
\begin{equation}\label{eq:ssd2}
    \mathbf{Z}_{n}^\text{grid} = \text{ReLU}(\text{Conv2d}(\mathbf{F}_{n}^\text{grid})), \mathbf{Z}_{n}^\text{grid} \in \mathbb{R}^{B \times C^\prime_d \times H_d \times W_d},
\end{equation}
where $C^\prime_d$ is the channel dimensionality of the backbone's output. These enriched features are shared among four task-specific decoding heads, each optimized for a distinct semantic task of image reconstruction, distance prediction, angle estimation, and velocity estimation.

Secondly, for the task of image reconstruction, a ViT decoder \cite{he2022masked} serves as the primary decoder. $\mathbf{Z}_{n}^\text{grid}$ undergoes up-sampling to double its spatial resolution, followed by the addition of a positional embedding, \(\mathbf{E}_\text{pos} \in \mathbb{R}^{H_d^\prime \times W_d^\prime \times D}\). This embedding retains spatial information during transformer-based processing:  
\begin{equation}\label{eq:ssd3}
    \mathbf{Z}^\text{embed}_{n} = \text{Upsample}(\mathbf{Z}_{n}^\text{grid}) + \mathbf{E}_\text{pos}, \mathbf{Z}^\text{embed}_{n} \in \mathbb{R}^{B \times H_d^\prime \times W_d^\prime \times D},
\end{equation}
where $\text{Upsample}(\cdot)$ is the upsampling operation. The transformer blocks within the ViT decoder perform global feature aggregation, and the final prediction layer maps the transformed features back to pixel space. The reconstructed image is derived via:  
\begin{equation}\label{eq:ssd4}
    \mathbf{\hat{m}}_{n} = \text{Unpatchify}(\text{Transformer}(\mathbf{Z}^\text{embed}_{n})),
\end{equation}
where $\text{Transformer}(\cdot)$ represents multiple transformer blocks and $\text{Unpatchify}(\cdot)$ converts a patch to an image.

Thirdly, the other sensing results, including angle, velocity, and distance, are extracted from the shared backbone features $\mathbf{Z}_{n}^\text{grid}$ through their specialized heads, respectively. This can be expressed as:  
\begin{equation}\label{eq:ssd5}
   \hat{\theta}_{n} = \operatorname{sigmoid}(\mathbf{W}_\text{angle}\cdot\text{Conv2d}(\mathbf{Z}_{n}^\text{grid})+\mathbf{b}_\text{angle}),
\end{equation}
\begin{equation}\label{eq:ssd6}
   \hat{v}_{n} = \operatorname{sigmoid}(\mathbf{W}_\text{rate}\cdot\text{Conv2d}(\mathbf{Z}_{n}^\text{grid})+\mathbf{b}_\text{rate}),
\end{equation}
\begin{equation}\label{eq:ssd7}
   \hat{d}_{n} = \operatorname{sigmoid}(\mathbf{W}_\text{distance}\cdot\text{Conv2d}(\mathbf{Z}_{n}^\text{grid})+\mathbf{b}_\text{distance}),
\end{equation}
where $\operatorname{sigmoid}(\cdot)$ is the sigmoid activation function, $\mathbf{W}_\text{angle}$, $\mathbf{W}_\text{rate}$, and $\mathbf{W}_\text{distance}$ are the weights of the three task output heads. $\mathbf{b}_\text{angle}$, $\mathbf{b}_\text{rate}$, and $\mathbf{b}_\text{distance}$ are the bias. 

Finally, we summarize the inference process of SSD in \textbf{Algorithm \ref{alg:SSD}}. The SSD efficiently integrates ViT-based image reconstruction and auxiliary sensing tasks. By optimizing a unified multitask objective, SSD can provide diversified sensing results using the same semantic information. Moreover, users can select suitable output heads to deploy locally, according to their personalized requirements.
\begin{algorithm}
\caption{Inference of SSD}
\label{alg:SSD}
\begin{algorithmic}[1]
    \REQUIRE $\mathbf{\hat{e}}_{n}$.
    \ENSURE $\mathbf{\hat{m}}_{n}, \hat{\theta}_{n}, \hat{v}_{n}, \hat{d}_{n}$.

    \STATE{Obtain the refined feature $\mathbf{Z}_{n}^\text{gird}$ using Eq. (\ref{eq:ssd2}).}
    \STATE{Reconstruct the image of the ST using Eqs. (\ref{eq:ssd3})-(\ref{eq:ssd4}).}
    \STATE{Predict the distance of the ST using Eq. (\ref{eq:ssd5}).}
    \STATE{Estimate the velocity of the ST using Eq. (\ref{eq:ssd6}).}
    \STATE{Predict the angle of the ST using Eq. (\ref{eq:ssd7}).}
\end{algorithmic}
\end{algorithm}

\subsection{Multi-Task Learning-Based Training Process}
To ensure the SIMAC framework can achieve diversified sensing services, multi-task learning is used to jointly train the modules in this framework.
Multi-task learning leverages a unified framework to address the simultaneous optimization of multiple objectives, targeting image reconstruction, angle estimation, and distance prediction. Specifically, the image reconstruction task is guided by the L1 loss, promoting pixel-level accuracy. Simultaneously, the angle, velocity, and distance prediction tasks are supervised using the mean squared error (MSE) loss, which minimizes deviations from the ground truth. The total multi-task loss is formulated as a weighted sum:
\begin{equation}\label{eq:MTL}
    \mathcal{L}_{\text{MTL}} = l_1\mathcal{L}_{\text{sr}} + l_2\mathcal{L}_{\text{ap}} + l_3\mathcal{L}_{\text{vp}} + l_4\mathcal{L}_{\text{dp}},
\end{equation}
where each task loss is given in Eqs. (\ref{eq:loss1})-(\ref{eq:loss4}).

During training, input data comprising the captured image $\mathbf{m}$ and echo signal $\mathbf{A}_{n}$. Moreover, dynamic channel characteristics are introduced based on the SNR and modulation schemes (e.g., BPSK, QPSK, 8PSK, 16QAM). These parameters, along with contextual information about the channel, are used to condition the model predictions. The forward pass of the model generates reconstructed images, estimated angles, and predicted distances. Losses for each task are computed and backpropagated to update the model parameters. 

Assuming the training dataset is $\mathcal{D}$, the training process of the SIMAC framework is summarized in \textbf{Algorithm \ref{alg:MTL}}. 
\begin{algorithm}
\caption{Training process based on multi-task learning}
\label{alg:MTL}
\begin{algorithmic}[1]
    \REQUIRE $\mathcal{D}$.
    \ENSURE $\bm{\alpha}, \bm{\beta}, \bm{\gamma}, \bm{\delta}, \bm{\epsilon}$.
    \FOR{each training epoch}
        \FOR{each batch $(\mathbf{m}_n, \mathbf{A}_{n})$ from $\mathcal{D}$ }
            \STATE{Generate communication parameters $C_n$ using dynamic SNR and modulation scheme.}
            \STATE{Predict the sensing results $\mathbf{\hat{m}}_{n}, \hat{d}_{n}, \hat{v}_{n}, \hat{\theta}_{n}$ according to \textbf{Algorithms \ref{alg:MSF}-\ref{alg:SSD}}.}
            \STATE{Compute total loss using Eq. (\ref{eq:MTL}).}
            \STATE{Backpropagate and update model parameters $\bm{\alpha}, \bm{\beta}, \bm{\gamma}, \bm{\delta}, \bm{\epsilon}$ with the optimizer.}
        \ENDFOR
    \ENDFOR
\end{algorithmic}
\end{algorithm}

\section{Experimental Setup and Numerical Results}
This section presents the simulation dataset, parameter configurations, and evaluation results. The simulations are conducted on a server equipped with an Intel Xeon CPU (2.3 GHz, 256 GB RAM) and two NVIDIA RTX 4090 GPUs (24 GB SGRAM each), leveraging the PyTorch framework to implement the proposed schemes. 

\subsection{Experimental Settings}
\subsubsection{Dataset Setup}  
Based on the VIRAT Video Dataset \cite{oh2011large}, we perform a series of operations to construct a specialized dataset for training and testing our proposed methods. The detailed procedure is as follows:  

First, we select videos representing three specific scenes from \cite{oh2011large} as the raw data. Each video is sampled at a velocity of one frame per second, resulting in approximately 10,000 RGB images (i.e., $\mathbf{m}$).  

Second, for each extracted frame, we assume that all STs are cars. A YOLOv10 \cite{wang2024yolov10} model detects the bounding box coordinates of cars in the scenes. Based on these bounding box coordinates, we employ the segment anything model (SAM) \cite{kirillov2023segment} to isolate the car images from the raw frames, which serve as labels for image reconstruction (i.e., $\mathbf{m}_{n}$). This process produces approximately 800,000 images of STs.  

Third, we assume the BS is positioned at the lower-right corner of each image. Accordingly, we calculate the distance of each ST to the BS as follows:  
\begin{equation}  
    d_{n}=\sqrt{(x^\text{BS}-x_{n}^\text{center})^2+(y^\text{BS}-y_{n}^\text{center})^2},  
\end{equation}  
where $(x^\text{BS}, y^\text{BS})$ represents the coordinates of the BS, and $(x_{n}^\text{center}, y_{n}^\text{center})$ denotes the center coordinates of the $n$th ST's bounding box.  

Similarly, we estimate the angle of each ST relative to the BS as follows:  
\begin{equation}  
    \theta_{n}=\operatorname{arctan2}(y^\text{BS}-y_{n}^\text{center}, x_{n}^\text{center}-x^\text{BS}),  
\end{equation}  
where $\operatorname{arctan2}(\cdot)$ computes the arc tangent of the input coordinates, returning a radian value within the range $[-\pi, \pi]$.  

Additionally, the tracking capability of the YOLO model is employed to track the same object across consecutive frames. The velocity of the object is calculated as the ratio of the displacement of the center point of the object detection bounding box to the frame duration. This can be expressed as:  
\begin{equation}
    v_n = \frac{\|\mathbf{p}_{n,t+1} - \mathbf{p}_{n,t}\|}{\Delta t},
\end{equation}
where \(\mathbf{p}_{n,t} = (x_{n,t}^\text{center}, y_{n,t}^\text{center})\) and \(\mathbf{p}_{n,t+1} = (x_{n,t+1}^\text{center}, y_{n,t+1}^\text{center})\) represent the center points of the $n$th ST at frame \(t\) and \(t+1\), respectively, and \(\Delta t\) is the time interval between the two frames.  

Finally, we generate the echo signal $\mathbf{A}_{n}$ for each ST according to Eq. (\ref{eq:sig}).

\subsubsection{Parameter Settings}  
In the system model, since we just consider a simple end-to-end communication, the AWGN channel is exclusively considered in the SC model. Considering the task of image reconstruction is more difficult than the other tasks,
the adjustment factors of task losses are set to $l_1=100$, $l_2=1$, $l_3=1$, and $l_4=1$, respectively.
The bandwidth is set to \( B = 1 \) kHz, the power is set to $P=1$ W, and the SNR is varied between 0 dB and 25 dB. When the SNR is low, the transmitted symbols are easy to be impacted. Hence, to achieve more accurate transmission, we apply the following modulation scheme:  
\begin{equation}
\text{Modulator} =  
	\begin{cases}  
		\text{BPSK}, & \text{0 dB} \leq \text{SNR} \leq \text{10 dB}, \\  
		\text{QPSK}, & \text{10 dB} < \text{SNR} \leq \text{18 dB}, \\  
		\text{8PSK}, & \text{18 dB} < \text{SNR} \leq \text{22 dB}, \\  
        \text{16QAM}, & \text{otherwise}. \\  
	\end{cases}  
\end{equation}  

The radar's central frequency is \( f_c = 10 \, \text{GHz} \). The PRI is configured as \( T_r = 1 \times 10^{-4} \, \text{s} \), with a sampling frequency of \( F_s = 60 \, \text{MHz} \), corresponding to a sampling interval of \( \Delta t = \frac{1}{F_s} = 1.67 \, \text{ns} \). Since only cars are considered as the ST, all RCSs of the STs are set to $\rho_n=100$.
We assume the SIMO radar model utilizes \( K = 10 \) antennas to transmit the LFM waveform and capture the echo signal.

During training, the SNR is randomly chosen for each forward pass to enhance the robustness of the SIMAC framework to channel noise. In the inference phase, we evaluate the SC model under fixed SNR conditions at $[0, 10, 15, 20, 25]$ dBs. 

\subsubsection{Benchmark Schemes}  
To evaluate the proposed SIMAC framework, we consider the following benchmark schemes:  
\begin{itemize}  
    \item SIMAC (w/o LSE): This variant excludes communication parameters during training and inference.  
    \item SIMAC (w/o SSD): In this variant, the SIMAC framework operates without multiple output heads, and each sensing task is trained independently.  
    \item DeepJSCC \cite{zhang2023predictive}: An image SC model incorporating predictive and adaptive deep coding, serving as an additional benchmark for the image reconstruction evaluation parts.  
\end{itemize}  

\subsubsection{Evaluation Metrics}
Firstly, we use the root mean squared error (RMSE) to evaluate the performance of the proposed method in distance, velocity, and angle prediction tasks. RMSE quantifies the absolute average deviation between predictions and ground truth. These metrics are defined as follows:  
\begin{equation}  
    \text{RMSE}(\mathbf{x}_i,\hat{\mathbf{x}}_i) = \sqrt{\frac{1}{I} \sum_{i=1}^I \| \hat{\mathbf{x}}_i - \mathbf{x}_i \|^2},  
\end{equation}  
where \( \hat{\mathbf{x}}_i \in (0, 1)\) denotes the predicted value, \( \mathbf{x}_i \) represents the ground truth, and \( I \) is the total number of samples. Particularly, $\mathbf{x}_i$ is the normalized value of speed, distance, or angle.

Secondly, to evaluate image reconstruction performance, we adopt peak signal-to-noise ratio (PSNR) and structural similarity index measure (SSIM) as metrics. PSNR measures the quality of reconstructed images and is expressed in decibels (dB), with higher values indicating better quality:  
\begin{equation}  
	\text{PSNR}(\mathbf{x}_i,\hat{\mathbf{x}}_i) = 10 \cdot \log_{10} \left( \frac{\text{MAX}_I^2}{\| \hat{\mathbf{x}}_i - \mathbf{x}_i \|^2} \right),  
\end{equation}  
where \( \text{MAX}_I \) is the maximum possible pixel value, typically 255 for 8-bit images.  
Similarly, SSIM is a metric that gauges the perceived similarity between two images, factoring in three key components - luminance, contrast, and structure. The definition of SSIM is outlined as follows:
\begin{equation}  
	\text{SSIM}(\mathbf{x}_i,\hat{\mathbf{x}}_i) = \frac{(2\varphi_{\mathbf{x}_i}\varphi_{\mathbf{\hat{x}}_i} + c_1)(2\phi_{\mathbf{x}_i\mathbf{\hat{x}}_i} + c_2)}{(\varphi_{\mathbf{x}_i}^2 + \varphi_{\mathbf{\hat{x}}_i}^2 + c_1)(\phi_{\mathbf{x}_i}^2 + \phi_{\mathbf{\hat{x}}_i}^2 + c_2)},  
\end{equation}  
where \( \varphi_{\mathbf{x}_i} \) and \( \varphi_{\mathbf{\hat{x}}_i} \) are mean values, \( \phi_{\mathbf{x}_i}^2 \) and \( \phi_{\mathbf{\hat{x}}_i}^2 \) are variances, \( \phi_{\mathbf{x}_i\mathbf{\hat{x}}_i} \) is their covariance, and \( c_1 \) and \( c_2 \) are two small constants to prevent division by zero.  

Lastly, since the reconstructed sensing image may be incorrect, we define the accuracy of the sensing as follows:
\begin{equation}
    \text{Accuracy} = \frac{N_\text{true}}{N},
\end{equation}
\begin{equation}
    N_\text{true} = \text{Count}(\{\mathbf{m}_n|\text{PSNR}(\mathbf{m}_n,\mathbf{\hat{m}}_n) \geq \text{15 dB}\}),
\end{equation}
where $\text{Count}(\cdot)$ is the calculation function.

\subsection{Evaluation Results}
This subsection aims to display the proposed SIMAC framework's running results and performance for several tasks, including distance, angle, velocity prediction, and image reconstruction.

\subsubsection{Visualization of Sensing Results}
\begin{figure*}[htbp]
	\centering
	\includegraphics[width=16cm]{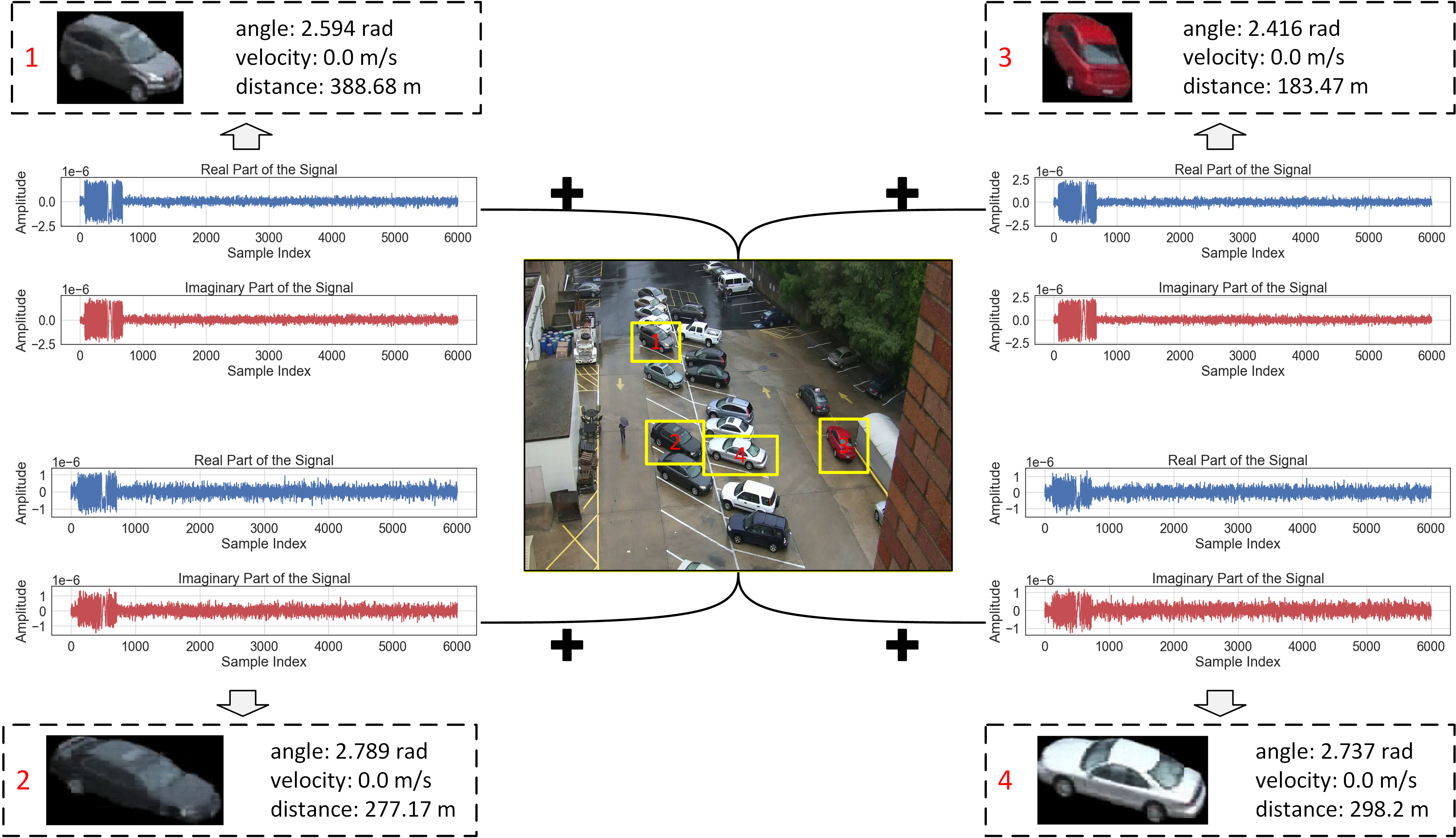}
	\caption{Visualization of the SIMAC framework's running process.}
	\label{fig:exp1}
\end{figure*}
\begin{figure*}[htbp]
	\centering
	\includegraphics[width=16cm]{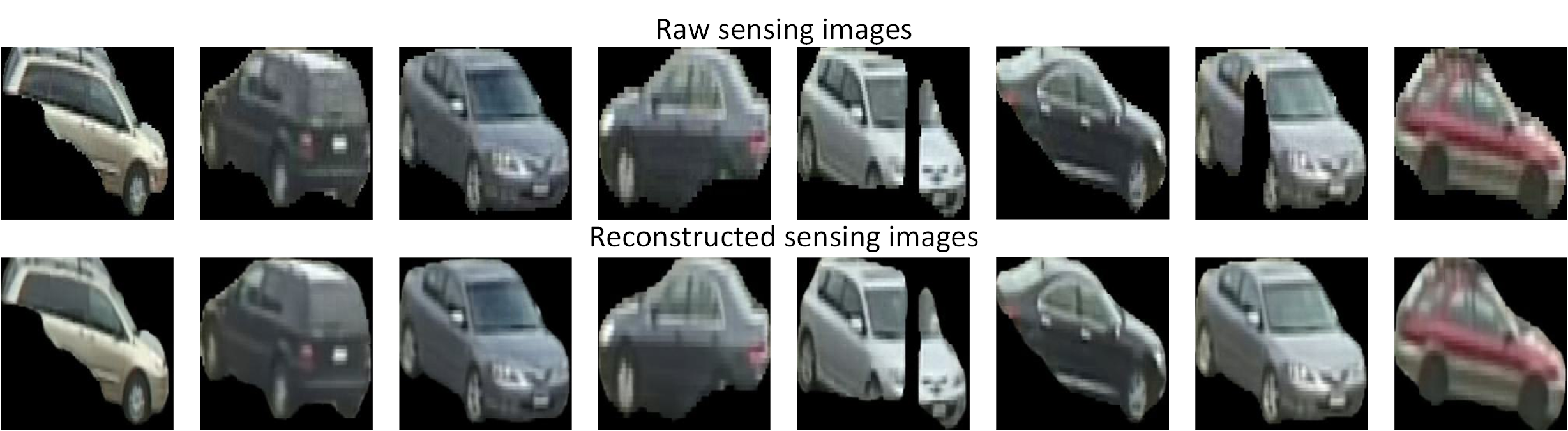}
	\caption{Comparison results of raw sensing images and reconstructed images.}
	\label{fig:compare}
\end{figure*}
\begin{figure}[htbp]
	\centering
	\includegraphics[width=8cm]{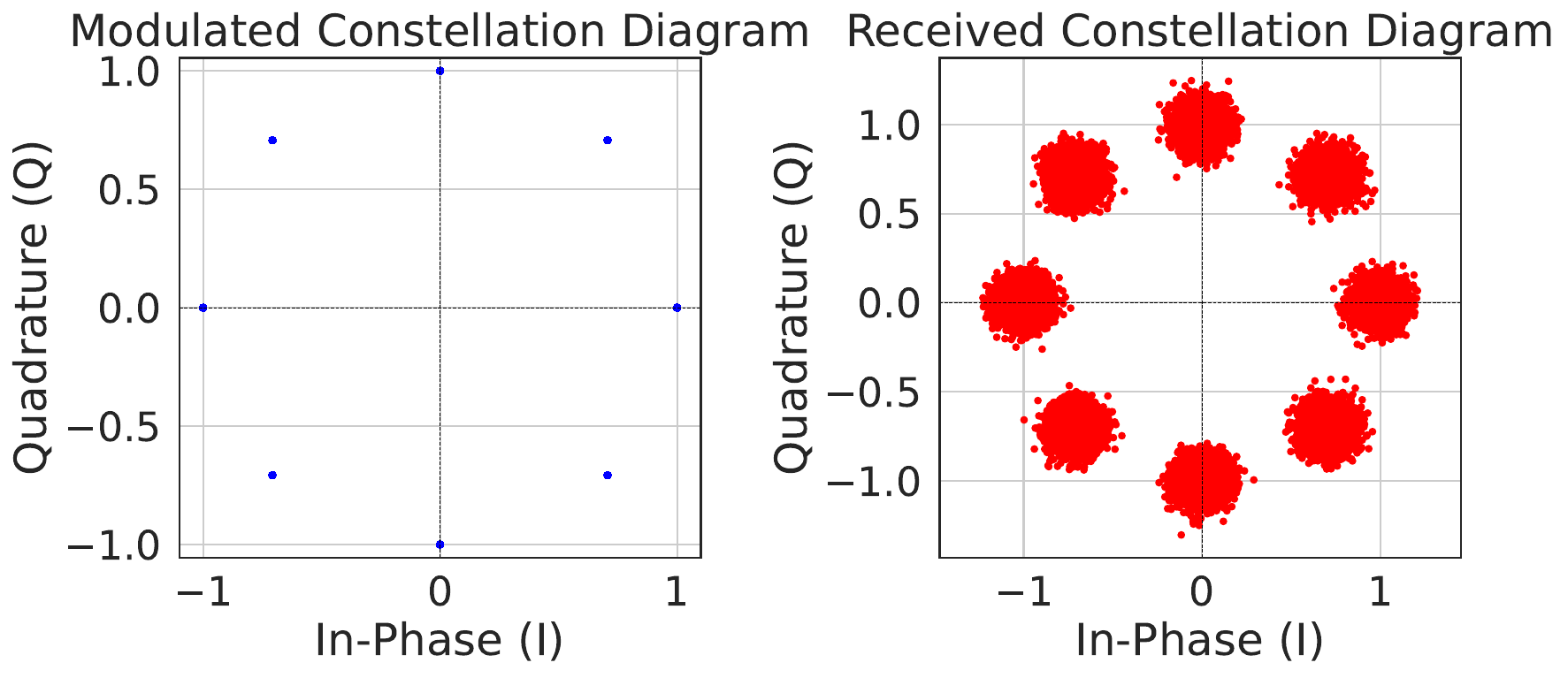}
	\caption{Constellation of the SIMAC when using 8PSK.}
	\label{fig:exp_8PSK}
\end{figure}
\begin{figure}[htbp]
	\centering
	\includegraphics[width=8cm]{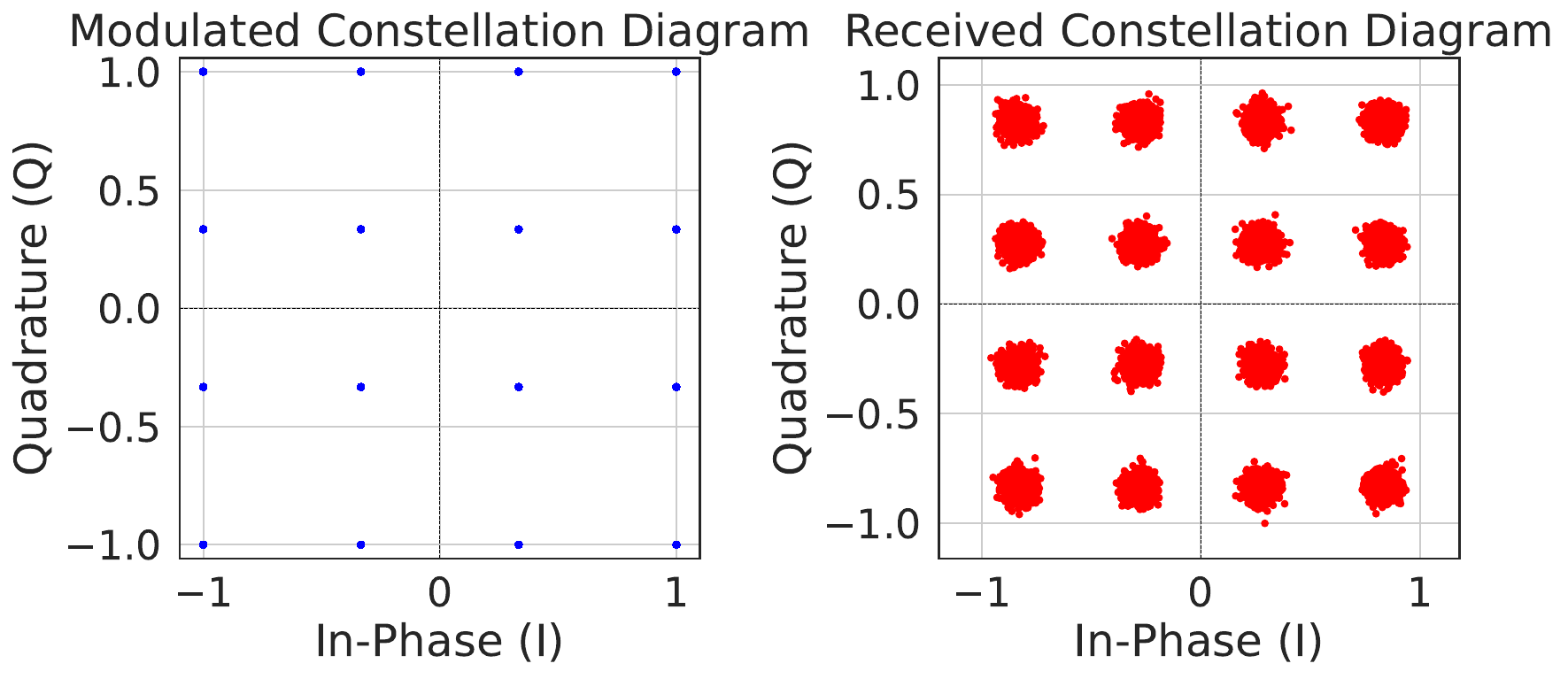}
	\caption{Constellation of the SIMAC when using 16QAM.}
	\label{fig:exp_16QAM}
\end{figure}
\begin{figure}[htbp]
	\centering
	\includegraphics[width=8.5cm]{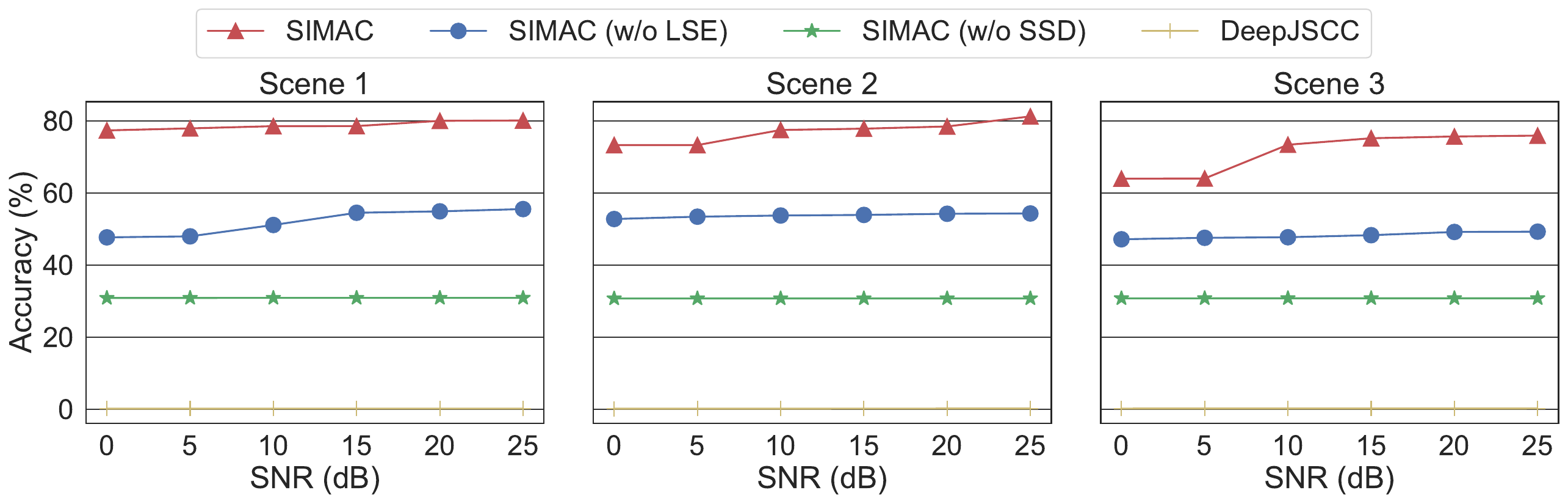}
	\caption{Accuracy comparisons across different schemes.}
	\label{fig:exp_acc}
\end{figure}
As illustrated in Fig. \ref{fig:exp1}, the SIMAC framework processes the raw image \(\mathbf{m}\), the radar signal \(\mathbf{A}_{n}\), and the sensing outputs, including the reconstructed image \(\mathbf{\hat{m}}_n\) and the predicted motion attributes. A distinctive feature of the SIMAC framework is its capability to produce diversified sensing outputs for the same image by integrating various radar signals.  
Since the training data is derived from simulations, there may be discrepancies between the predicted results and real-world scenes.  
Fig. \ref{fig:compare} presents a partial comparison between the raw sensing images and the reconstructed images. The reconstructed images exhibit superior visual quality and are capable of restoring missing parts of the original images. This can be attributed to the robust generative capabilities of the ViT decoder.
Figs. \ref{fig:exp_8PSK} and \ref{fig:exp_16QAM} present the constellations for 8PSK and 16QAM modulation schemes, respectively, demonstrating the framework’s resilience to channel noise. Furthermore, Fig. \ref{fig:exp_acc} compares the sensing accuracy across three distinct scenes (Scene 1, Scene 2, and Scene 3), underscoring SIMAC's superior performance. In contrast, the SIMAC (w/o SSD) and DeepJSCC schemes show the lowest accuracy. This can be attributed to the absence of signal modality assistance in these methods, which hinders their ability to achieve precise visual positioning.

Overall, the proposed MSF module enables the SIMAC framework to leverage the semantic information of radar signals, which serve as queries for localizing corresponding spatiotemporal features during training. These findings demonstrate the framework’s capacity to fuse multimodal information effectively for target localization and motion attribute estimation.  

\subsubsection{Evaluation for Distance Sensing}
\begin{figure}[htbp]
	\centering
	\includegraphics[width=8.5cm]{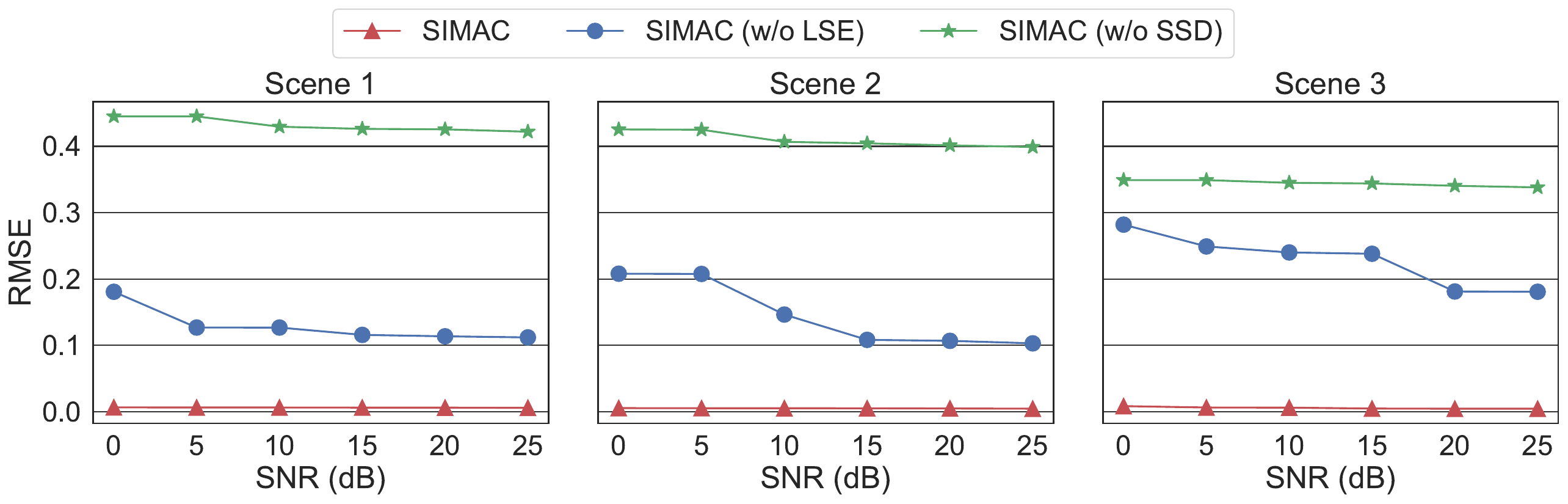}
	\caption{Distance prediction comparisons across different schemes.}
	\label{fig:exp_dis}
\end{figure}

Fig. \ref{fig:exp_dis} presents the RMSE of distance predictions across three distinct scenes under varying SNR levels. The SIMAC framework consistently delivers the lowest RMSE, nearly zero across all SNR levels and scenes, demonstrating its accuracy and robustness.  
The SIMAC (w/o LSE) exhibits a gradual performance decline as noise increases, underscoring the critical role of the LSE module in mitigating noise effects. The SIMAC (w/o SSD) scheme achieves the worst RMSE performance, emphasizing the SSD module’s importance to precise distance predictions.  

\subsubsection{Evaluation for Velocity Sensing}
\begin{figure}[htbp]
	\centering
	\includegraphics[width=8.5cm]{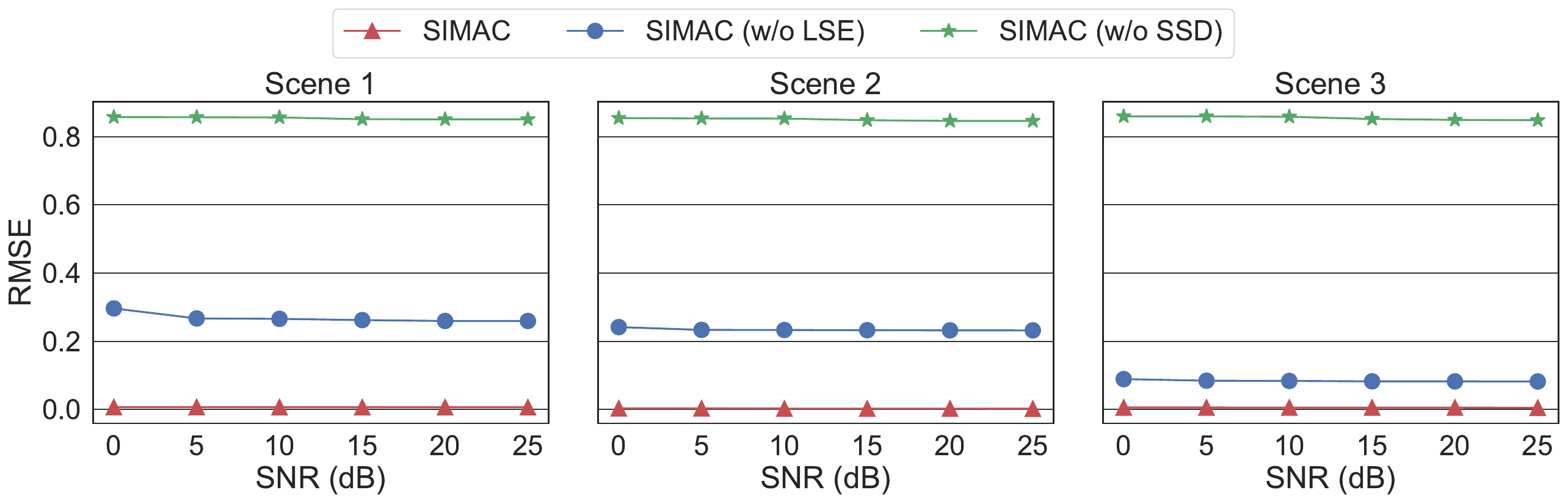}
	\caption{Velocity prediction comparisons across different schemes.}
	\label{fig:exp_rate}
\end{figure}
Fig. \ref{fig:exp_rate} illustrates the velocity sensing performance of the different schemes, where the SIMAC framework achieves the best RMSE across all SNR levels and scenes. 
The SIMAC (w/o LSE) shows moderate performance with an RMSE of around 0.2 across all conditions, indicating that removing LSE compromises accuracy. The SIMAC (w/o SSD) exhibits the worst RMSE, stable at approximately 0.8, reflecting significant performance degradation in the absence of the SSD module. 
This reflects the effectiveness of the full SIMAC framework and the critical roles of both LSE and SSD components in achieving precise velocity predictions.

\subsubsection{Evaluation for Angle Sensing}
\begin{figure}[htbp]
	\centering
	\includegraphics[width=8.5cm]{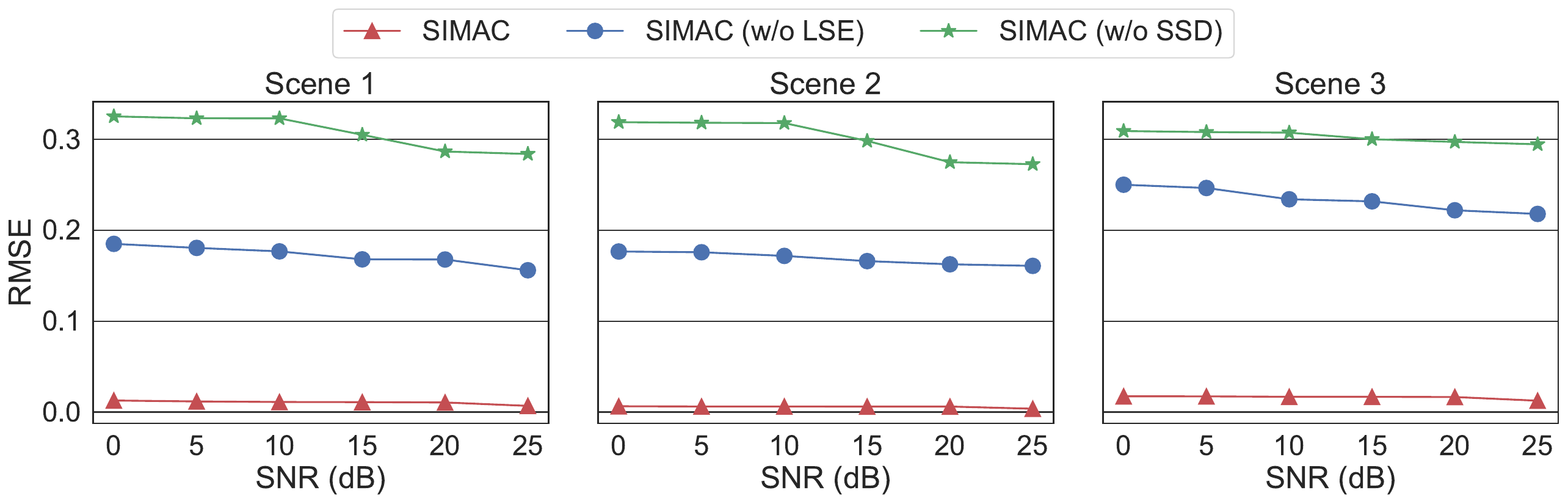}
	\caption{Angle prediction comparisons across different schemes.}
	\label{fig:exp_angle}
\end{figure}
Fig. \ref{fig:exp_angle} displays the angle sensing performance of the different schemes,  where the SIMAC scheme achieves the best RMSE within all the SNR levels and scenes. In contrast, SIMAC (w/o SSD) exhibits the worst RMSE, remaining stable at approximately 0.3, suggesting significant performance degradation without the SSD component. 
Meanwhile, SIMAC (w/o LSE) shows a moderate RMSE, consistently around 0.2 across all SNR levels, indicating a noticeable decline in accuracy due to the absence of LSE. 
These results demonstrate the superiority of the complete SIMAC framework and the essential contributions of both LSE and SSD components in achieving precise velocity predictions across diversified scenes.

\subsubsection{Evaluation for Image Reconstruction}
\begin{figure}[htbp]
	\centering
	\includegraphics[width=8.5cm]{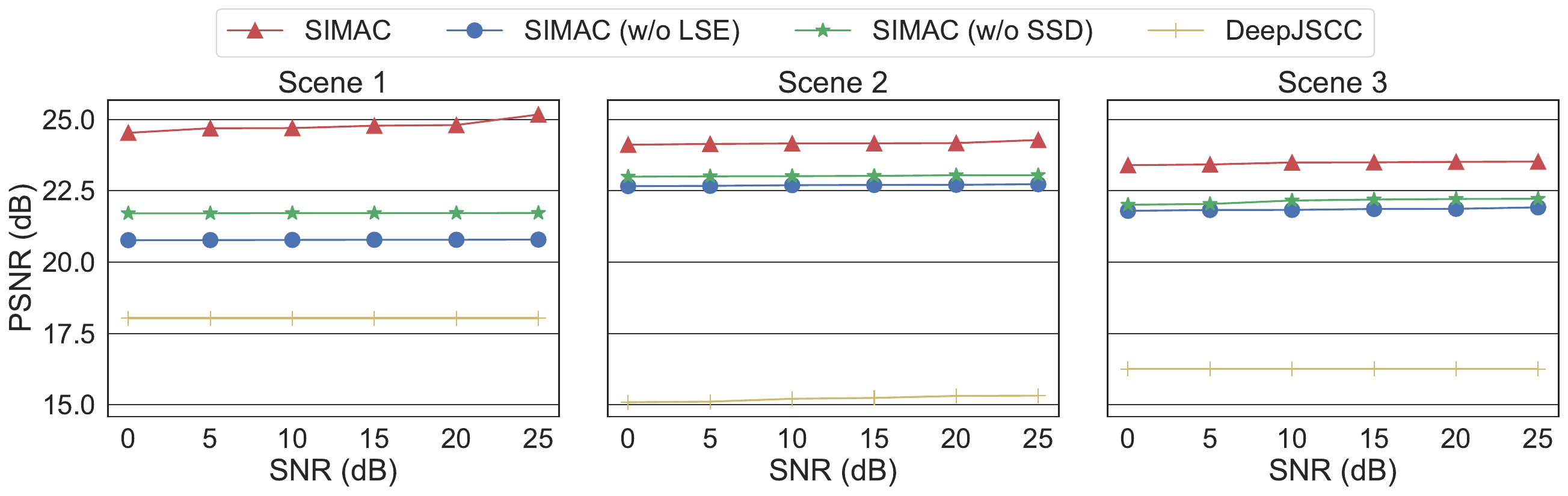}
	\caption{PSNR comparisons across different schemes.}
	\label{fig:exp_psnr}
\end{figure}
\begin{figure}[htbp]
	\centering
	\includegraphics[width=8.5cm]{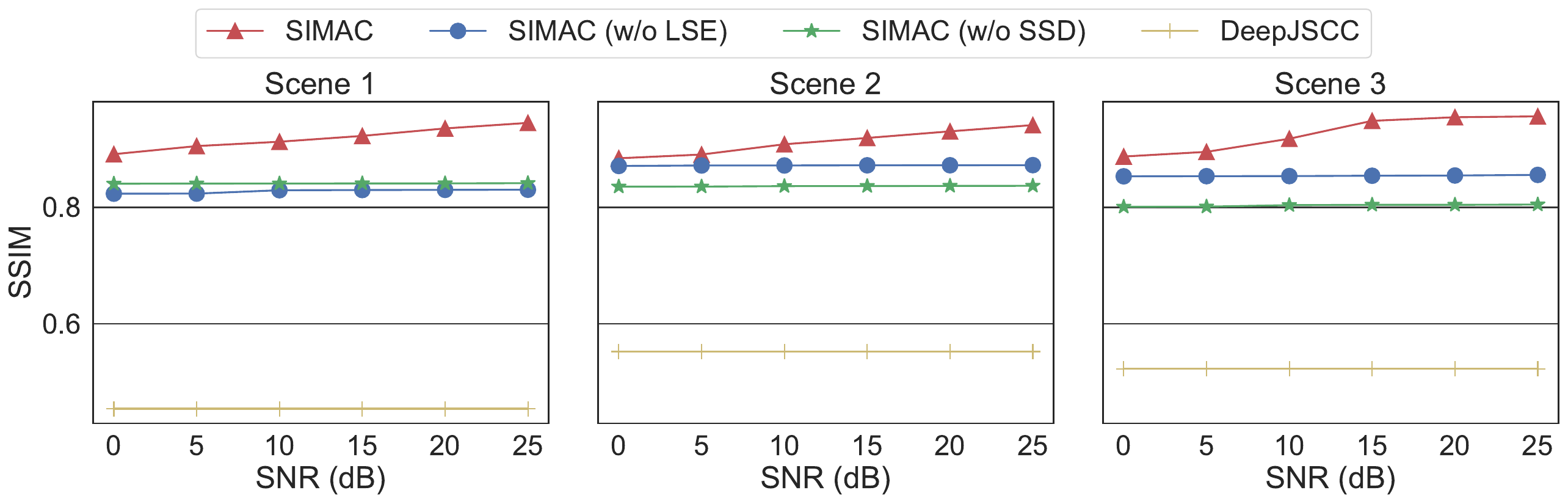}
	\caption{SSIM comparisons across different schemes.}
	\label{fig:exp_ssim}
\end{figure}

Fig. \ref{fig:exp_psnr} shows the image reconstruction performance of the different schemes in the three scenes in terms of PSNR.  In Scene 1, the SIMAC scheme demonstrates superior performance, with a PSNR consistently exceeding 24 dB, as well as showing a slight increase as SNR levels rise. In contrast, SIMAC without LSE exhibits significantly lower PSNR values, remaining stable around 21 dB across all SNR levels. Similarly, SIMAC without SSD shows moderate performance, achieving PSNR values around 22 dB, indicating limited improvement over SIMAC without LSE. For Scene 2 and Scene 3, SIMAC continues to outperform the other schemes, though the performance gap narrows slightly as SNR increases.  DeepJSCC, in contrast, performs significantly worse across all scenes, with PSNR values consistently below 18 dB, regardless of SNR. This starkly highlights the limitations of DeepJSCC, which relies solely on unimodal visual data. The absence of multi-modal integration makes it challenging for DeepJSCC to effectively reconstruct the target, particularly under channel noise conditions.  
Fig. \ref{fig:exp_ssim} illustrates the SSIM performance across different schemes, presenting a trend similar to the PSNR of Fig. \ref{fig:exp_psnr}. The SIMAC framework achieves the highest SSIM scores, consistently exceeding 0.85 in Scene 1, with slight improvements as SNR increases. SIMAC without LSE and SIMAC without SSD exhibit lower SSIM values, while DeepJSCC remains the worst performer, with SSIM values consistently below 0.7, due to its reliance on unimodal data.  

These results underscore the robustness and effectiveness of the complete SIMAC framework in maintaining high reconstruction quality. Both LSE and SSD play critical roles in optimizing PSNR and SSIM performance across varying conditions. Furthermore, the poor performance of DeepJSCC emphasizes the necessity of multi-modal data integration for accurate target localization and robust sensing performance.  

\section{Conclusions and Future Works}
To address the challenges of limited accuracy and restricted capabilities in single-modality sensing, high communication overhead in decoupled sensing-communication systems, and the inability of single-task sensing to meet diversified user demands, we propose the SIMAC framework. This framework integrates multimodal sensing with SC to enable low-cost and high-accuracy sensing services. 
Specifically, the framework first employs the MSF network to extract and fuse semantic information from radar signals and images using cross-attention mechanisms, generating comprehensive multimodal representations. Then, it incorporates the LSE that maps communication parameters and multimodal semantics into a unified latent space, enabling channel-adaptive semantic encoding. Furthermore, it introduces the SSD to feature task-specific decoding heads and a multi-task learning strategy to deliver diversified sensing services. 
We conducted experimental simulations across four sensing tasks and benchmarks, demonstrating that the SIMAC framework substantially enhances sensing accuracy and service diversity.

In future work, we aim to extend the proposed framework by incorporating multi-base station collaborative sensing to enhance the robustness and scalability of the system. By fusing the sensing results across multiple base stations from different perspectives, it is possible to achieve a three-dimensional reconstruction of sensing targets, providing richer spatial information and more comprehensive sensing capabilities for applications such as autonomous driving and smart surveillance.

\bibliographystyle{IEEEtran}
\bibliography{bare_jrnl}
\newpage
\end{document}